\documentclass[letterpaper, 10 pt, conference]{ieeeconf}  

\IEEEoverridecommandlockouts                              

\usepackage{graphicx} 
\usepackage[caption=false, font=footnotesize]{subfig}
\usepackage{xcolor}
\usepackage{soul}
\usepackage{textcomp, gensymb}
\usepackage{svg}
\usepackage{diagbox}

\usepackage{multirow}
\usepackage{caption}
\newcommand{\cornerlabel}[0]{\captionsetup[subfigure]{justification=raggedright, singlelinecheck=false, skip=0mm, captionskip=-4mm, margin={1mm,0cm}}}

\title{\LARGE \bf
Budget-optimal multi-robot layout design for box sorting
}

\author{Peiyu Zeng$^{1*}$, Yijiang Huang$^{2*}$, Simon Huber$^2$, Stelian Coros$^2$
\thanks{$^{1}$ Peiyu Zeng is with the Department of Mechanical and Process Engineering, ETH,
Zurich, Switzerland. {\tt\small pezeng@ethz.ch}}%
\thanks{$^{2}$ The authors are with the Department of Computer Science, ETH,
Zurich, Switzerland. 
{\tt\small \{yijiang.huang; simon.huber; scoros\}@inf.ethz.ch}}%
\thanks{$^*$ These authors contributed equally.}%
}

\usepackage[utf8]{inputenc} 
\usepackage[T1]{fontenc}
\usepackage{graphicx}
\usepackage{amsmath}
\usepackage{amssymb}

\usepackage{csvsimple}
\usepackage{adjustbox}
\usepackage{longtable}
\usepackage{booktabs}
\usepackage{wrapfig}

\usepackage{float}

\usepackage{placeins}

\usepackage[shortcuts]{extdash}

\usepackage{csquotes}
\usepackage{algorithm}
\usepackage[noend]{algpseudocode}

\definecolor{junglegreen}{rgb}{0.16, 0.67, 0.53}

\usepackage{hyperref}
\hypersetup{
    colorlinks=true,
    citecolor=teal,
    linkcolor=teal, 
    filecolor=magenta,      
    urlcolor=cyan,
}

\usepackage[capitalize]{cleveref}
\creflabelformat{equation}{#2\textup{#1}#3}
\Crefname{figure}{Fig.}{Fig.}

\newcommand{\x}[0]{\mathbf x}
\newcommand{\y}[0]{\mathbf y}

\begin{document}

\maketitle
\thispagestyle{empty}
\pagestyle{empty}

\begin{abstract}
Robotic systems are routinely used in the logistics industry to enhance operational efficiency, but the design of robot workspaces remains a complex and manual task, which limits the system's flexibility to changing demands.
This paper aims to automate robot workspace design by proposing a computational framework to generate a budget-minimizing layout by selectively placing stationary robots on a floor grid to sort packages from given input and output locations.
Finding a good layout that minimizes the hardware budget while ensuring motion feasibility is a challenging combinatorial problem with nonconvex motion constraints.
We propose a new optimization-based approach that models layout planning as a subgraph optimization problem subject to network flow constraints.
Our core insight is to abstract away motion constraints from the layout optimization by precomputing a kinematic reachability graph and then extract the optimal layout on this ground graph.
We validate the motion feasibility of our approach by proposing a simple task assignment and motion planning technique.
We benchmark our algorithm on problems with various grid resolutions and number of outputs and show improvements in memory efficiency over a heuristic search algorithm.
In addition, we demonstrate that our algorithm can be extended to handle various types of robot manipulators and conveyor belts, box payload constraints, and cost assignments. A supplementary video demonstrating the proposed framework and its results is available on YouTube at \url{https://www.youtube.com/watch?v=W8xWlnaLUxA}.
\end{abstract}
\section{Introduction}
Robotic systems are routinely used in the logistics industry to enhance operational efficiency, which often involves the use of conveyor belts and robotic arms to facilitate the sorting and transportation of packages. 
Regardless of the specific machinery employed, every logistics workflow requires a well-designed workspace that effectively meets operational demands.
However, designing such workspaces remains a complex and manual task, which limits the system's flexibility when adapting to changing demands, for example, a sudden influx of goods to a new destination that requires adding a new output port.
Our goal is to automate the workspace design to enable adaptive robotic sorting systems that can rapidly reconfigure their layout and function.
\begin{figure}[thpb]
    \centering
    \cornerlabel
    \subfloat[]{%
        \includegraphics[width = 1.0\linewidth]{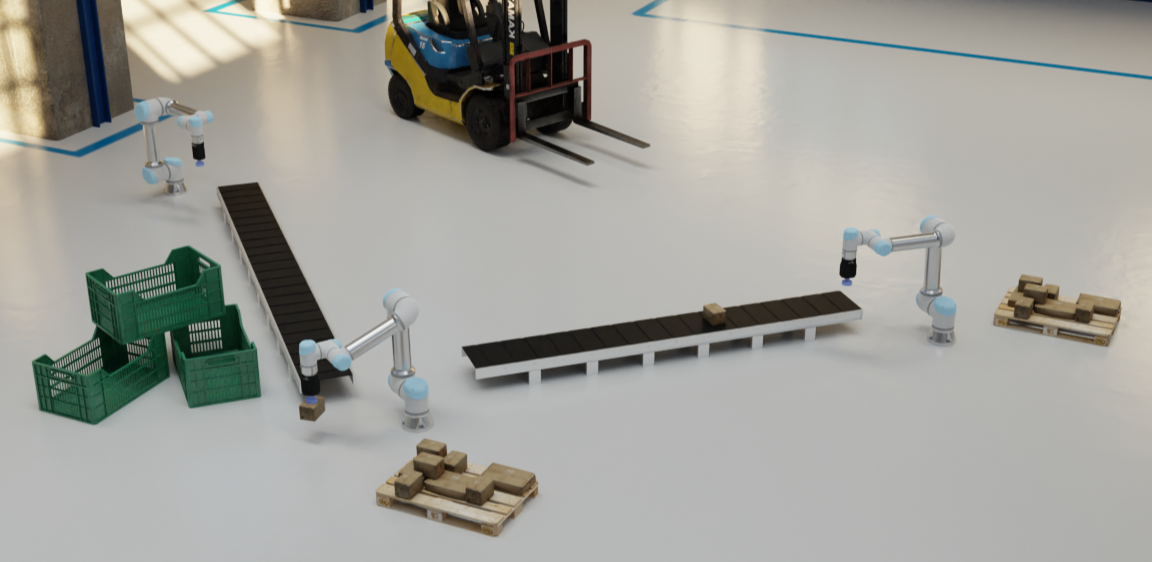}
    }\\
    \subfloat[]{%
        \includegraphics[width = 1.0\linewidth]{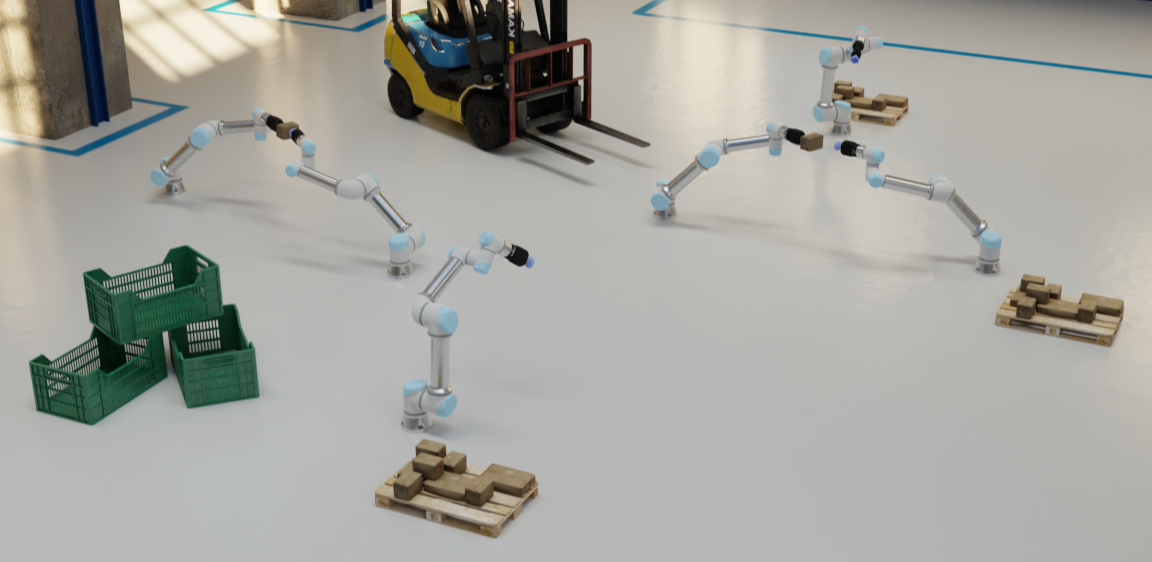}
    }
    \caption{Multi-robot logistic layouts generated using our approach: (a) before and (b) after adding an output location.}
    \label{fig:teaser}
\end{figure}

This work presents a computational framework to optimize a robot layout to minimize hardware budget while ensuring a feasible collaborative plan to deliver boxes from a given input to several output locations.
We focus on designing the fully actuated components of a workspace, i.e., the selection and placement of stationary robots on a floor grid, which include robotic arms and conveyor belts, and assume that the non-actuated components are provided and fixed, such as input ports and output containers.

Designing a multi-stationary robot workspace that minimizes the hardware budget requires solving a challenging optimization problem that involves discrete and continuous variables: one needs to decide {\it which} robots to use and {\it where} to put them on the floor, while ensuring valid robot motions exist to deliver boxes to targets.
Our key insight is to decouple the layout optimization from motion planning, and build a graph to model the flows of goods for each possible robot placement. 
Then, we formulate the layout design problem as finding a minimum-cost subgraph while ensuring flows to all destinations, which can be solved using Mixed-Integer Linear Programming (MILP).

Our core contributions include:
\begin{itemize}
    \item We propose a new formulation of the budget-optimal multi-robot layout problem as a minimum-cost, connected subgraph extraction problem. We solve our new formulation using mixed integer linear programming.
    \item Our formulation provides a versatile framework to incorporate different types of robot manipulators and conveyor belts, payload constraints, and cost assignments. 
    \item We validate our method on problems with various grid resolution and number of output ports. 
    As the first work solving the multi-robot layout problem, we compare our method to a heuristic search approach operating on the reachability graph, and show that our method is memory-efficient, up to 100 times faster, and more versatile for extension.
\end{itemize}

    
\section{Related Work}

\textbf{Layout optimization for robot manipulators}
involves placing robots within a workspace to maximize task efficiency and operational robustness. 
A key metric in this process is reachability, which evaluates a robot's ability to manipulate objects across different positions and orientations within its operational domain.
In \cite{Zacharias2007Capturing}, a directional reachability map is introduced to represent the manipulatable regions, which can be precomputed per robot on a voxelized space \cite{Makhal2018Reuleaux}.
A reachability score can be derived for optimizing the robot placement to ensure multi-directional manipulability \cite{Xu2021Optimal,wang2023multi}.
In these previous works, the layout only involves placing one or two robots working in close proximity, and the interaction between robots plays no role in the layout.
In contrast, our work considers placing robots on a large open floor where a successful delivery requires multi-agent collaborative manipulation, and the number of robots is to be decided in the layout optimization.

\textbf{Co-design of robots and trajectories}
considers co-optimizing a robot's design parameters (e.g., link lengths) with its motion trajectories or control policy, so that the robot's hardware can be tailored to the specific deployed environment, enhancing performance and reducing energy consumption.
Robot link lengths have been co-optimized with joint trajectories using sensitivity analysis \cite{ha2018computational} or inverse kinematic methods \cite{whitman2018task}, but typically assuming that the end-effector trajectories are provided.
\cite{Toussaint2021Co} embeds design parameter optimization into a sequential manipulation formulation to optimize the robot arm's morphology for a task ensemble. 
\cite{Huber2021Designing} co-optimizes links and motor configuration of animatronic figures with motion trajectories using a search with a custom-designed heuristic.
While previous work focuses on a single robot's "intrinsic" design parameters, such as link lengths, our work focuses on designing the layout of multiple robots, which could be regarded as "extrinsic" properties of an agent in a larger system. 

\textbf{Multi-agent path finding coupled with layout design}
has been investigated in the context of robotic warehouses. 
\cite{ZhangIJCAI23,zhang2024arbitrarily} propose learning-based methods to optimize the layout to maximize the throughput of 2D mobile robots in a warehouse using a MAPF algorithm as the performance evaluator. 
\cite{gao2023environment,gao2023constrained} consider the environment layout as decision variables, jointly optimizing it with agent performance and environment cost through reinforcement learning to achieve high performance in MAPF.
Although their problem is similar to ours on a conceptual level, they focus on routing paths for mobile robots on a 2D grid. In contrast, our work focuses on finding the minimum-cost layout for robotic manipulators.

\section{Problem Overview}

Our goal is to determine the layout of a multi-robot delivery system that can send boxes from a single input collection point to \(N\) destinations on a factory floor.
We assume that the robots can only be installed on a discrete grid of points \(P\subset\mathbb{R}^2\).
This setting is commonly used in reconfigurable factories, where a regular grid of installation anchors are pre-drilled into the floor.
We can assign robots from a given set of available robot types $\mathcal{R}$, without a limit placed on the number of each type.

At the layout planning phase, we only know about the high-level demands, i.e., the input/ouput port configurations, but not the detailed influx timeline, which is typically only known at deploy time.
Thus, we consider the delivery of each box separately, i.e., boxes are not flowing in with a specific order, and thus focus on the geometric feasibility of the system to deliver boxes.
The decision variables include:

\begin{itemize}
    \item \textit{Robot layout assignment}: let $a: P \rightarrow \mathcal{R} \cup \{Nil\} $ be the assignment function that associates a grid point $p$ with an appropriate robot type in R. $Nil$ means no robot is assigned;
    \item \textit{Task schedules}: for delivering box \(i\), let $s^i: [0,T] \rightarrow \mathcal{R}$ be the scheduling function that maps time $t$ to robots that are actively grasping box $i$.
    \item \textit{Trajectory} of each robot: \(\mathbf{x}_{r_j}: [0,T] \rightarrow \mathbb{R}^{d_j}\), where $d_j$ is the degree of freedom of robot $r_j$.
\end{itemize}

Among these, \textit{Robot layout assignment} is the primary variable of interest of our work, while the other two variables serve as auxiliary components to validate the motion feasibility.

Given the input and output locations of the box, the points of the grid, and the set of available robots, we aim to generate an optimal layout that minimizes the total hardware budget while ensuring feasible robot movements to deliver each box:
\begin{subequations}
    \begin{align}
        \min_{a,s,\x,\y} \ &\sum_{p\in P} \textsc{Cost}(a(p)) \\
        s.t. \quad &\forall i \in {1, \cdots, N}\nonumber\\
        &g_i(\x(t)) \leq 0, \forall t \in [0,T] \label{constr:ineq}\\
        & c_i(\y(t), \x_j(t)) = 0,\ \forall t,j\ s.t.\ r_j = s^i(t) \label{constr:grasp}\\
        & c^*_i(\y(t)) = 0, \forall t \in \{0,T\}\qquad\qquad \label{constr:initial}
    \end{align}
    \label{eq:naive_formulation}
\end{subequations}
where \(\y\) is the trajectory of all boxes, the \cref{constr:ineq} denotes general constraints that ensure the robot trajectory for delivering box $i$ is collision-free and respects joint limits. 
\Cref{constr:grasp} defines the grasping relationship between the box $i$ and the active robot manipulator $r_j$. 
\Cref{constr:initial} defines the initial and target poses of each box \(i\).

Directly solving \eqref{eq:naive_formulation} is intractable due to its combinatorial nature and the coupling between discrete layout assignment and continuous trajectory optimization. 
To tackle this, we adopt a hierarchical approach that uses a precomputed kinematic reachability graph (\cref{sec:precompute_graph}) to abstract away the motion constraints from the layout optimization.
This enables us to formulate layout planning as a minimum-cost, connected subgraph extraction problem, which can be solved to global optimality (\cref{sect:subgraph_extraction}).


\section{Method}


\subsection{Precomputing reachability graph}\label{sec:precompute_graph}

Formally, we aim to build a directed graph $G=(V,A)$, where the vertex set $V=V_\mathrm{io} \cup V_\mathcal{R}$ contains input/output locations $V_\mathrm{io}$ and all possible robot assignment $V_\mathcal{R}$.
The robot arrangement ground set $V_\mathcal{R}$ is constructed by filling each grid point $p \in P$ with all robots in $\mathcal{R}$, with multiple vertices occupying the same physical grid point.

\begin{figure}[b!]
    \centering
    \cornerlabel
    \subfloat[\label{fig:reachable1}]{%
        \includegraphics[height = 0.2\linewidth]{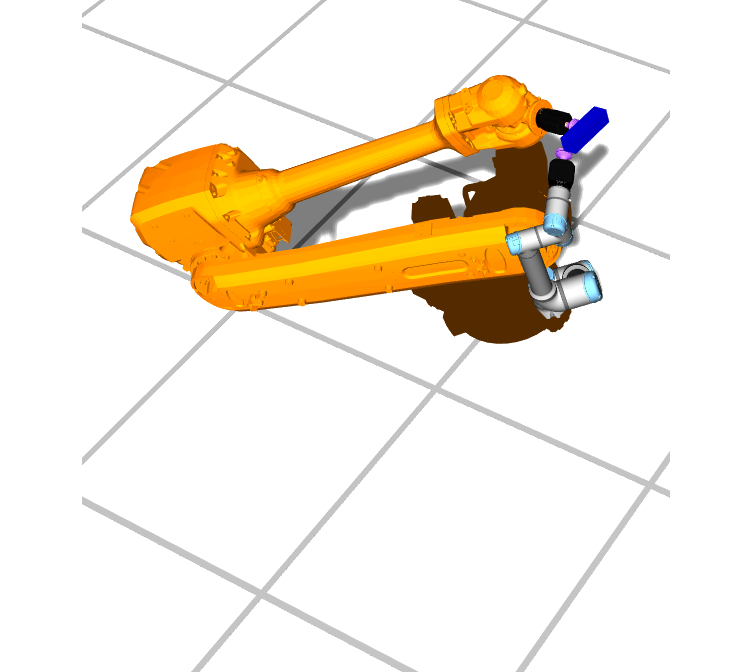}
    }
    \subfloat[\label{fig:reachablie3}]{%
        \includegraphics[height = 0.2\linewidth]{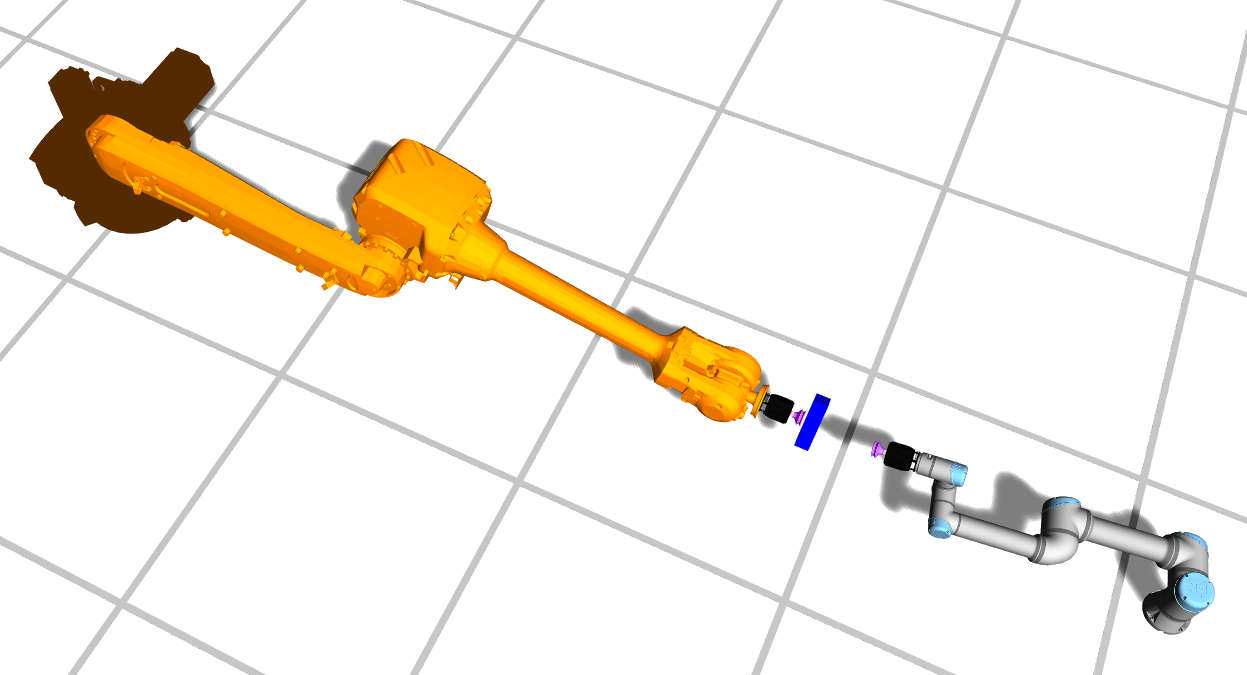}
    }
    \subfloat[\label{fig:reachablie2}]{%
        \includegraphics[height = 0.2\linewidth]{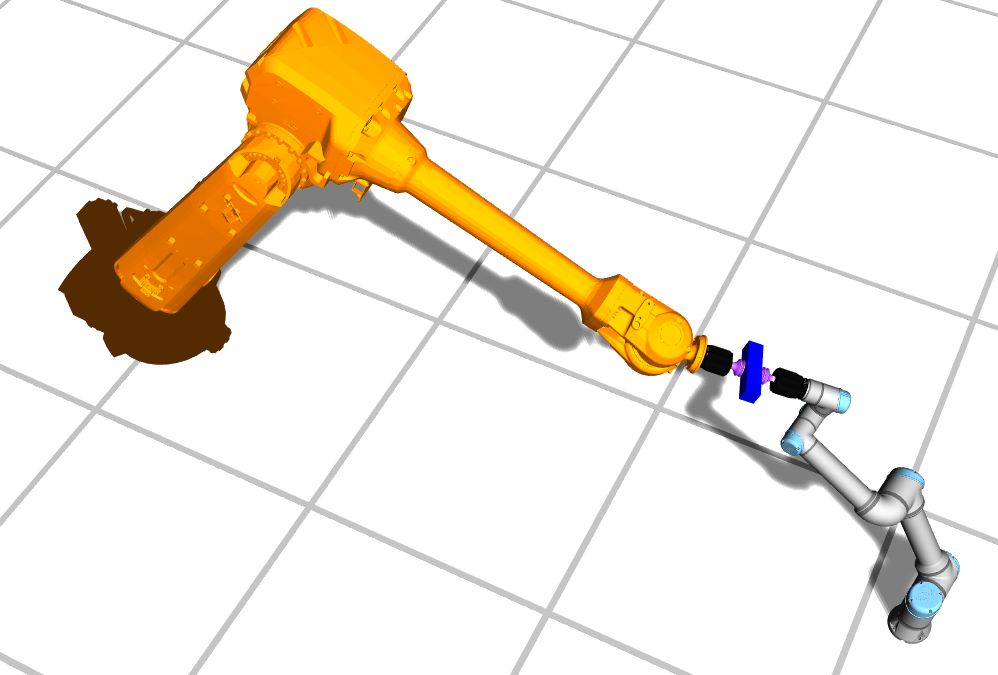}
    }
    \caption{Non-reachable robot pairs due to collisions (a) and kinematic limits (b), and a reachable robot pair (c)}
    \label{fig:reachable}
\end{figure}

\begin{figure}[thpb]
    \centering
    \cornerlabel
    \subfloat[\label{fig:layout}]{%
        \includegraphics[width = 0.45\linewidth]{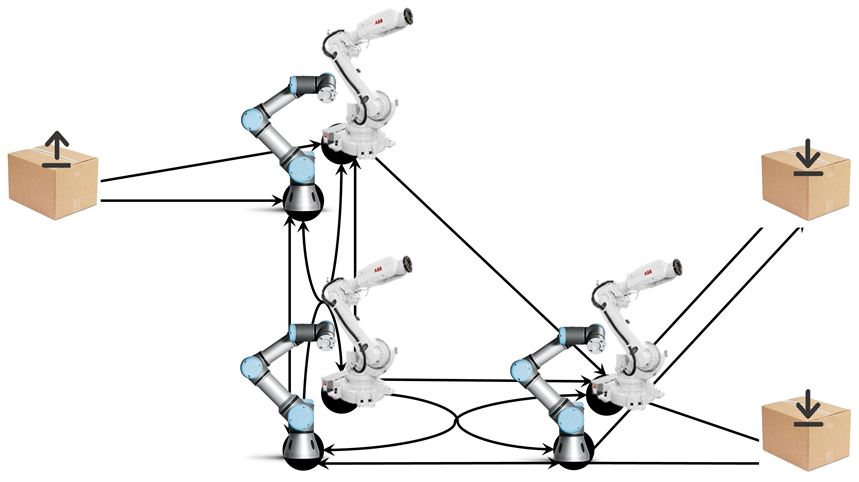}
    }
    \subfloat[\label{fig:subgraph}]{%
        \includegraphics[width = 0.45\linewidth]{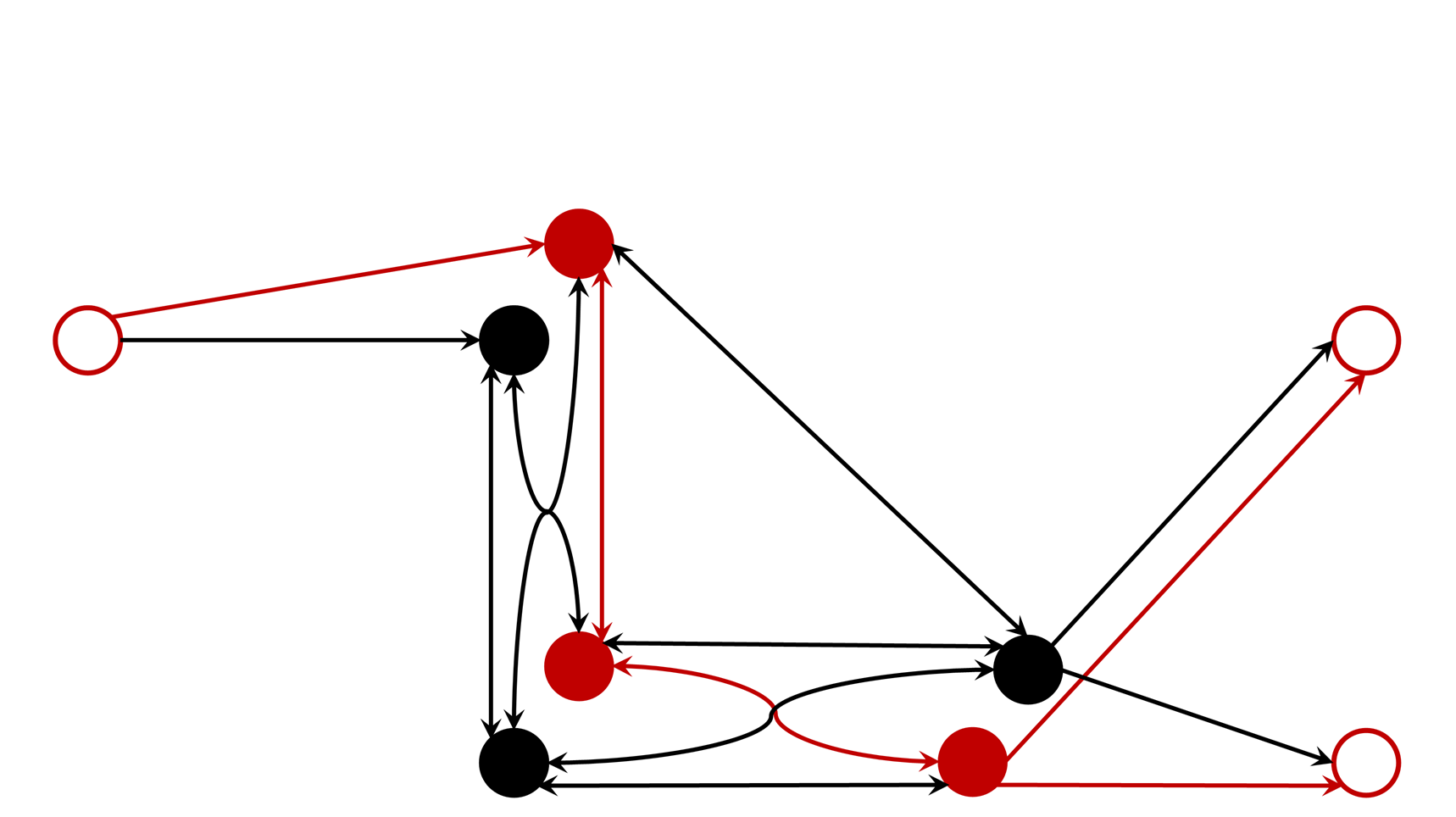}
    }
    \caption{The ground set and the corresponding graph (a) of the layout, where each grid point is assigned with two potential robots, and a feasible subgraph (b, red) extracted from it. We use hollow vertices to denote input/output locations and solid vertices to denote robots. }
\end{figure}

Each arc $a = (v_i, v_j)$ of the arc set $A$ indicates that a vertex $v_i$ can be reached by $v_j$ while satisfying the kinematic and collision constraints.
This is checked using an optimization-based inverse kinematics formulation with end-effector constraints for handover and sphere-based collision constraints \cite{zimmermann2020multi}.
The examples of the results of the reachability check are shown in \cref{fig:reachable}.
\Cref{fig:layout} shows a visual example of the robot layout ground set, and \cref{fig:subgraph} shows the corresponding directed graph.

\subsection{Subgraph extraction}\label{sect:subgraph_extraction}

Since all kinematic constraints are embedded in the reachability graph, the layout optimization can be simplified as extracting a minimum-cost subgraph, while ensuring that there exist paths to connect the input to all outputs.
An example feasible subgraph is colored red in \cref{fig:subgraph}.
By assigning the cost of robots as the corresponding vertex weights $w_v$, our goal becomes minimizing the total vertex weight in a subgraph where all output locations are connected to the input location with paths, which can be formulated as:
\begin{subequations}
    \begin{align}
        \min_{G' = (V', A')} \ &\sum_{v \in V'} w_v \label{eq:preliminary_objective}\\
        s.t.\ &V' \subseteq V, A' \subseteq A\label{eq:subgraph}\\
        &V_\mathrm{io}\subset V'\label{eq:vertices_inhertited}\\
        &\exists \psi = (v_1, v_2, \ldots, v_m)\ s.t.\ (v_i,v_{i+1})\in A',\notag\\
        &v_1 = v_\mathrm{i}, v_m = v_\mathrm{o},\forall v_\mathrm{i} \in V_\mathrm{in}, \forall v_\mathrm{o} \in V_\mathrm{out}\label{eq:connectivity}
    \end{align}
    \label{eq:preliminary}
\end{subequations}
%

\subsection{Solving with MILP}

\Cref{eq:preliminary} can be further formulated as a mixed-integer linear programming (MILP) problem using network flow.
The key idea is to transform vertex weights into arc weights and introduce binary arc selection variables to control the arcs' flow capacity.
We first substitute every robot vertex $v \in V_r$ with two vertices inheriting inbound arcs and outbound arcs, respectively, and an auxiliary arc $a$ connecting them with arc weight inherited from the original vertex: $w_a = w_v$.
Then, the original arc set $A$ is augmented with the set of auxiliary arcs $A_\mathrm{aux}$.

Let \(s_a\in \{0, 1\},\ a \in A\) be the binary variable indicating whether the arc \(a\) is selected in the subgraph. 
To model \cref{eq:connectivity}, we introduce \(N\) commodity flows, each starting from the input location and ending at one of the output locations.
Let \(f_{a,i},\ a \in A,\ i \in \{1,2,\ldots,N\}\) be the nonnegative flow value on the arc \(a\) for box \(i\). 
The optimization problem can be formulated as:
\begin{subequations}
    \label{eq:layout_optimization}
    \begin{align}
        \min_{\substack{
        s_a \in \{0,1\}, 
        \ f_{a,i}\in\mathbb{R}_{\ge 0}, \\
        a \in A,\ 
        i \in \{1,2,\ldots,N\}}}
        \sum_{a \in A_\mathrm{aux}} w_a s_a,
    \end{align}\label{eq:optimization_objective}
    \begin{align}
        s.t.&\sum_{\substack{a \in A_\mathrm{aux},\\
        \textsc{coor}(a) = p}} 
        s_a \le 1, \ \forall p \in P\label{eq:non_overlapping}\\
        &\forall i \in \{1,2,\ldots,N\},\nonumber\\
        &\qquad\quad\ f_{a, i} \le s_a,\:\qquad\qquad\qquad \forall a \in A \label{eq:capacity}\\
        &\ \begin{alignedat}{4}
            \sum_{a\in\delta^+(v)} &f_{a,i} - &\sum_{a\in\delta^-(v)} f_{a,i} &=&&\ -1, \forall v \in V_\mathrm{in},\\
            \sum_{a\in\delta^+(v)} &f_{a, i} - &\sum_{a\in\delta^-(v)} f_{a, i} &=&&\ 1, \ v = V_{\mathrm{out}, i},\\
            \sum_{a\in\delta^+(v)} &f_{a, i} - &\sum_{a\in\delta^-(v)} f_{a, i} &=&&\ 0,\ \mathrm{otherwise}
        \end{alignedat}
        \label{eq:flow_consistency}
    \end{align}
\end{subequations}
where $\textsc{coor}(a)$ maps an auxiliary arc $a$ to the location $p$ of the corresponding robot,
\(\delta^+(v)\)/\(\delta^-(v)\) represents the inbound/outbound arc set of vertex \(v\). 
\Cref{eq:non_overlapping} avoid multiple robots being installed at the same location in space.
\Cref{eq:capacity} restricts the flow value on all arcs not exceeding the capacity limits controlled by the selection variable. 
\Cref{eq:flow_consistency} guarantees that the influx equals the outflux on all vertices for all flows, which ensures the existence of a flow that connects each box's the input and output locations.

Such a flow problem can be solved with any off-the-shelf MILP solver, which typically implements a variation of the branch-and-bound algorithm with presolve, cutting planes, and domain-agnostic heuristics \cite{huang2021branch}.
When the solver converges, the solution is guaranteed to be optimal, upon which we can trace the optimal subgraph as the layout.



\subsection{Layout validation through task scheduling and motion planning} \label{sect:motion_planning}

The generated layout guarantees kinematic reachability to deliver each box, but it does not consider the temporal information of box input when a sequence of boxes are delivered asynchronously.
To demonstrate the validity of the coordinated delivery motions, we adopt a simple task assignment and motion planning workflow described below. 
This strategy is used to generate all the robot motions in the supplementary video.

For task scheduling, we need to determine the delivery order of all boxes, and determine the start and end time for each robot task.
To achieve shorter execution times for illustration purposes, we use a delivery order that prioritizes the longer graph transport path.
With the optimal subgraph representing the layout computed, the robot assignment \((r^i_1, r^i_2, ..., r^i_{N_i})\) for delivery box $i$ can be determined by tracing the shortest path connecting the box's input and output locations.
Assuming that every robot takes equal time $\Delta T$ to transport a box, the task schedule for box $i$ is $s^i([(k-1)\times\Delta T, k\times\Delta T]) = r^i_{k}, k \in \{1,\cdots, N_i\}$.
Then, we begin with having all tasks start at the same time and shift the conflicting tasks to ensure no robot is delivering two boxes at a time.
For motion planning, we adopt the optimization-based motion planning formulation described in \cite{zimmermann2020multi}.

Note that the delivery ordering and task assignment is heuristic and only to demonstrate that the layout admits a collision-free, asynchronized motion plan.
In a realistic situation where influx boxes emerge in an unpredictable timeline, dynamic task allocation \cite{harris2022fc} and reactive control strategies \cite{toussaint2022sequence} can be applied on top of our layout, and are left as future work.

\section{Extensions}\label{sect:extensions}

Due to the MILP formulation, our method is highly versatile in incorporating additional linear constraints, allowing extensions such as conveyor belt systems and robot arms with different payload capacities.

\subsection{Modeling conveyor belts}\label{sect:conveyor_belt}
\begin{figure}[thpb]
    \centering
    \subfloat[Inline junction\label{fig:inline_junction}]{%
        \includegraphics[width = 0.31\linewidth]{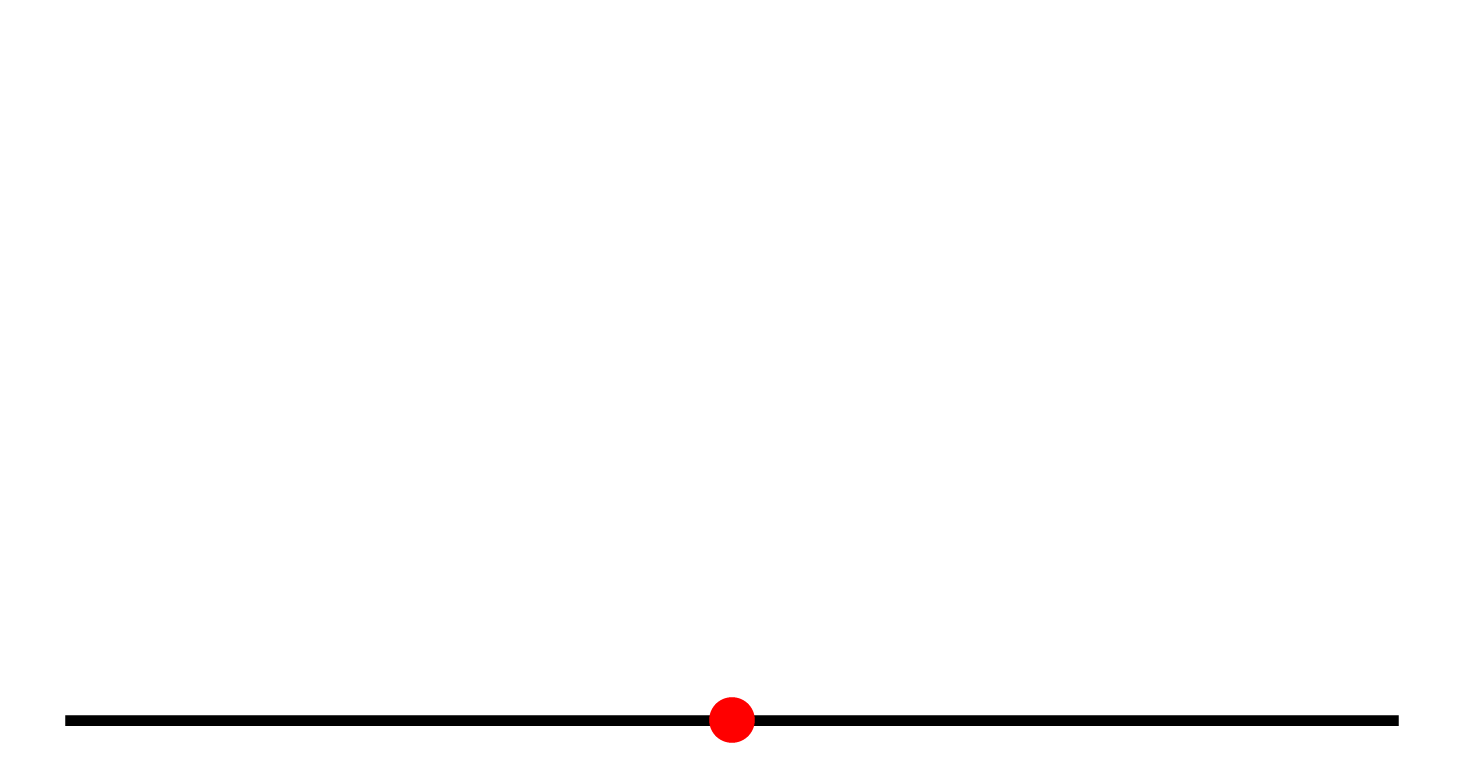}
    }
    \subfloat[Multi-way junction\label{fig:multi-way_junction}]{%
        \includegraphics[width = 0.31\linewidth]{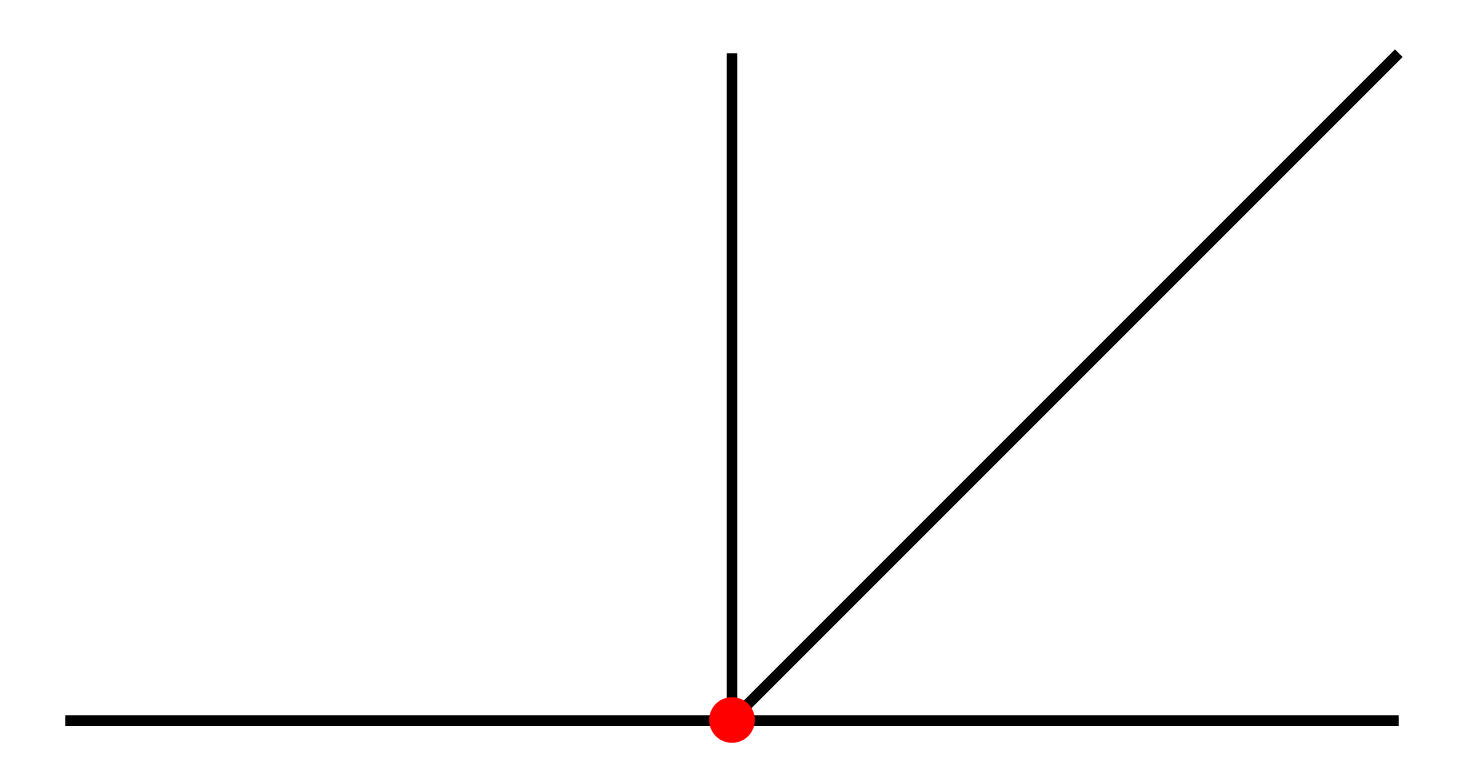}
    }
    \subfloat[Turning junction\label{fig:turning_junction}]{%
        \includegraphics[width = 0.31\linewidth]{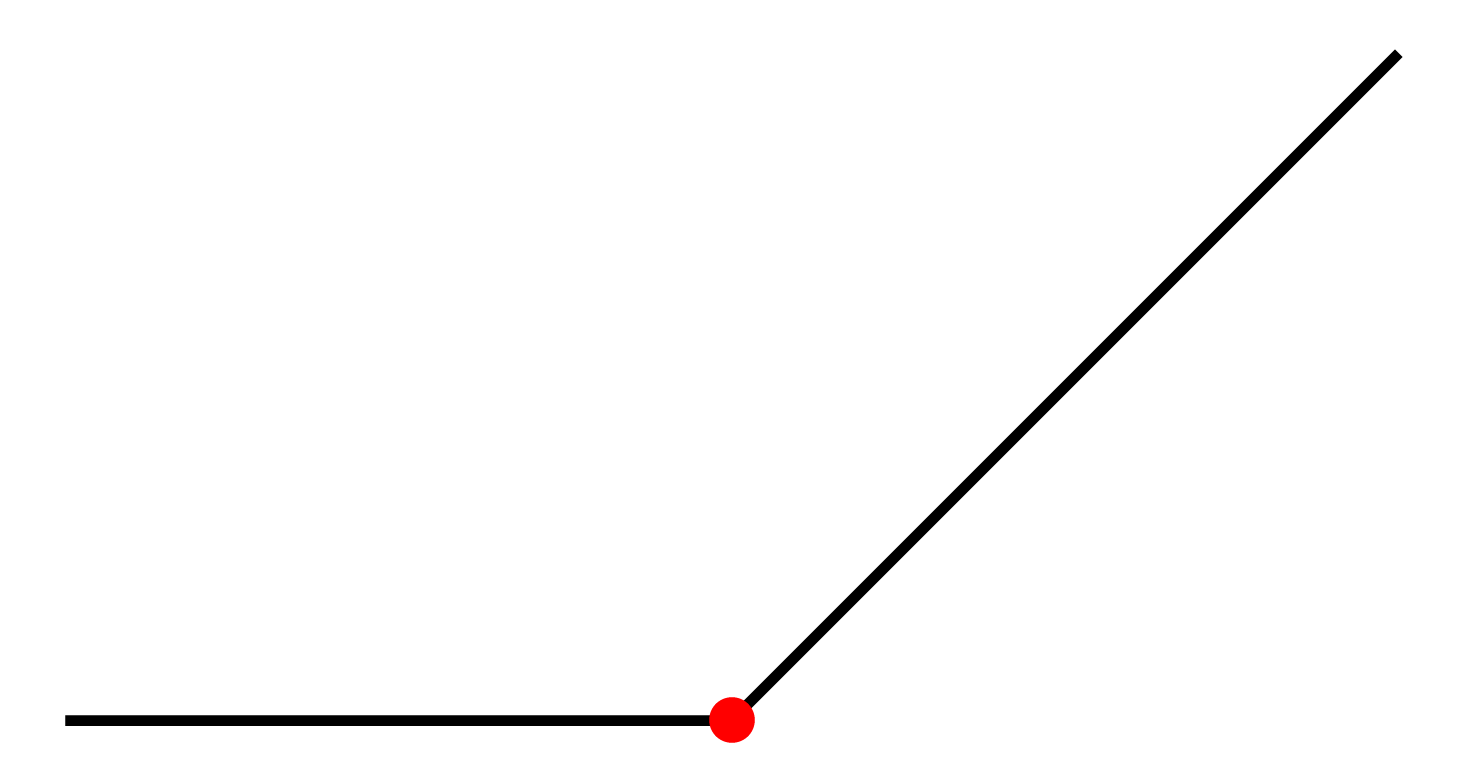}
    }
    \caption{Three types of junctions to connect belt segments}
    \label{fig:junctions}
\end{figure}

\begin{figure}[thpb]
    \centering
    \cornerlabel
    \subfloat[\label{fig:conveyor_belt_layout}]{%
        \includegraphics[width = 0.48\linewidth]{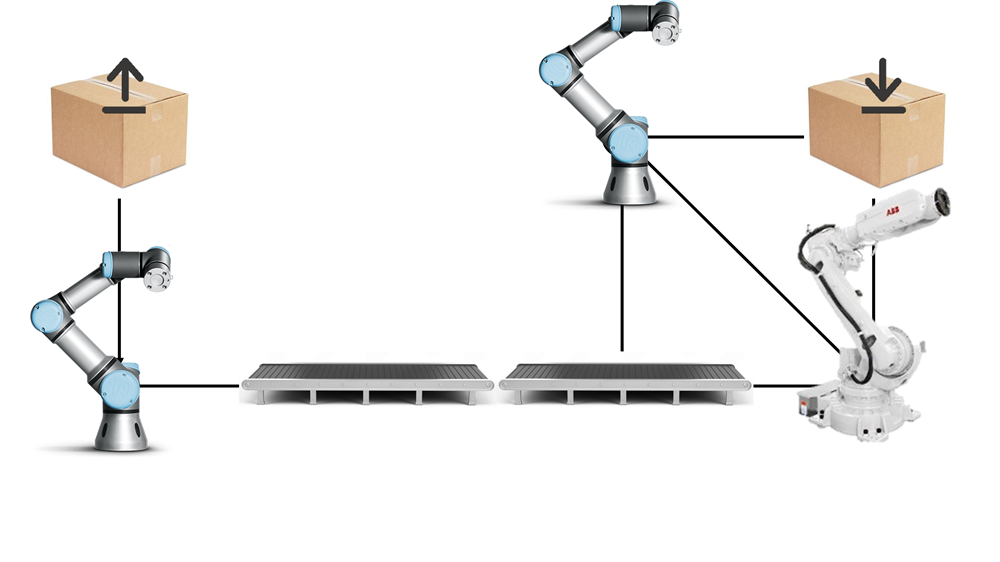}
    }
    \subfloat[\label{fig:conveyor_belt_augmented_graph}]{%
        \includegraphics[width = 0.48\linewidth]{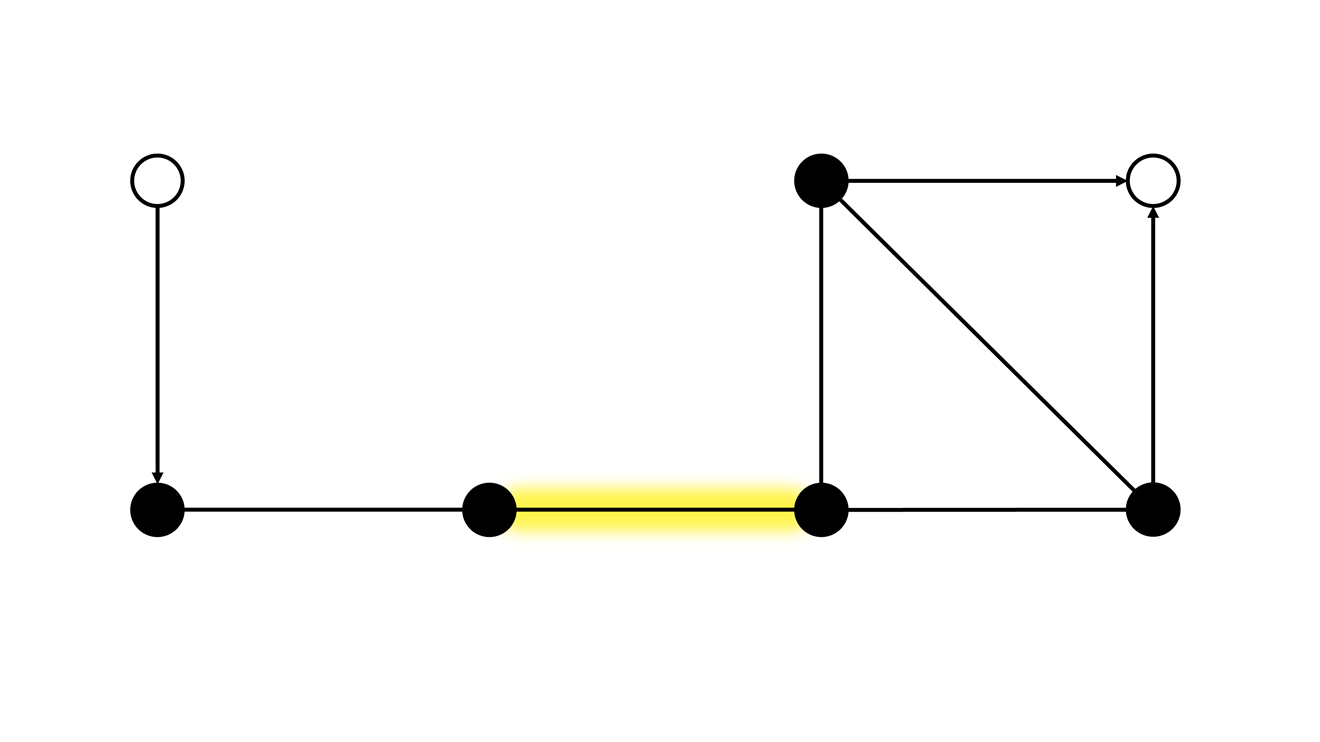}
    }
    \caption{A feasible layout with conveyor belts (a) and its corresponding graph representation (b), where the arc representing the inline junction is highlighted.}
    \label{fig:conveyor_belt_example}
\end{figure}

We use a modular conveyor belt system consisting of unit segments of different orientations, which can be mounted on the floor grid points.
For simplicity, we only allow four orientations: 0\degree, 45\degree, 90\degree, and 135\degree. 
Similar to the turntables in a real railway system, junctions are needed to connect conveyor belt segments and to direct boxes to different angles. 
We use three types of junctions as shown in \cref{fig:junctions}: 
{\it inline junctions} are fictitious links that merge two conveyor belt segments into a longer, straight one;
{\it turning junctions} redirect objects in a pair of pre-defined input/output directions;
{\it multi-way junctions} redirect objects to multiple directions.
\Cref{fig:conveyor_belt_example} shows an exemplary feasible layout with conveyor belts.


When constructing the reachability graph for the layout optimization, 
conveyor belt segments are treated similarly to robot arms.
However, junctions require special treatment because their selection depends on the selected segments in their neighborhood.
Below, we describe how each junction type is represented in the reachability graph, and the corresponding constraints to be added to the MILP formulation in \cref{eq:layout_optimization}.

\begin{figure}[thpb]
    \centering
    \cornerlabel
    \subfloat[\label{fig:inline_graph}]{%
        \includegraphics[height = 0.319\linewidth]{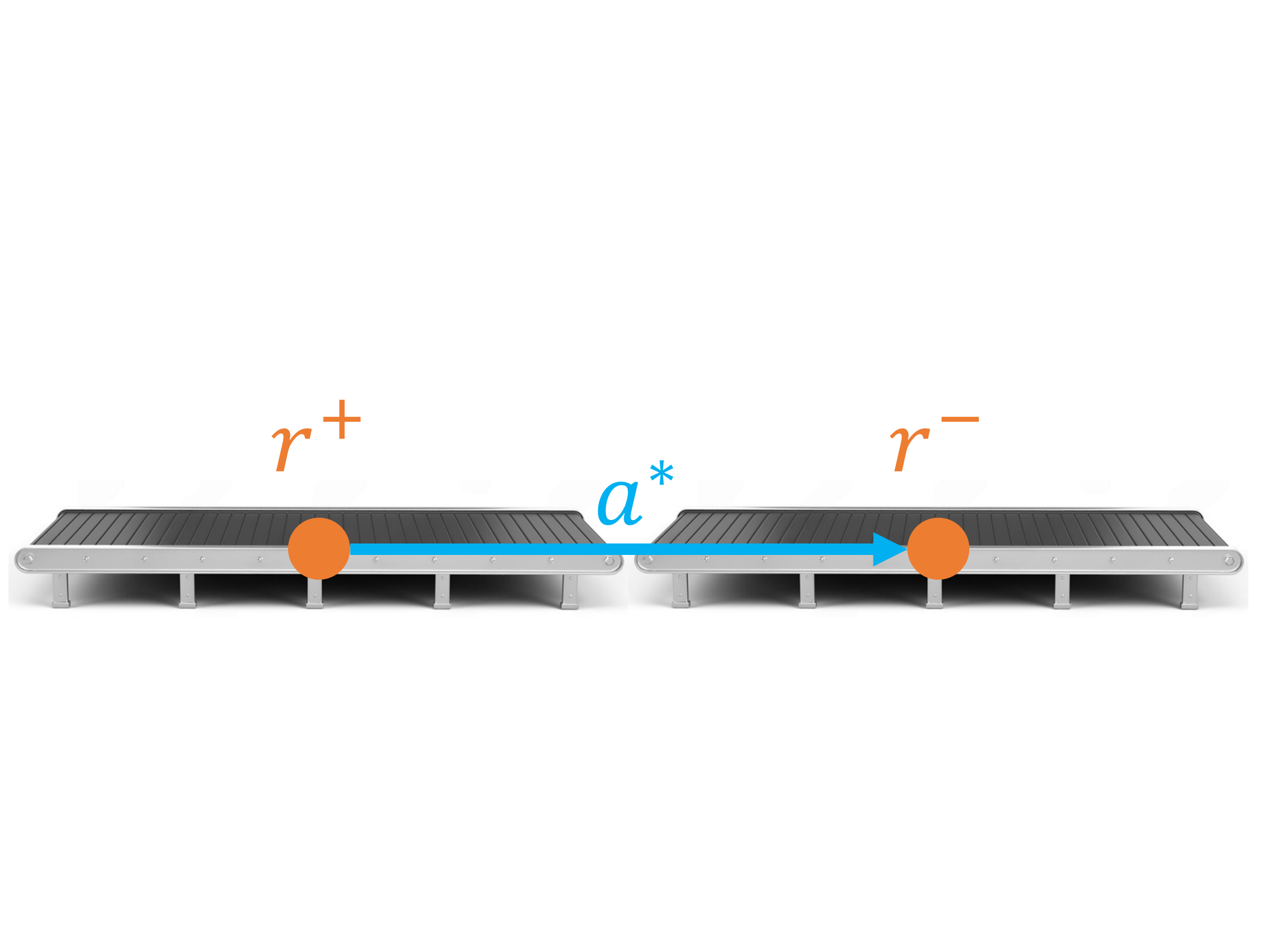}
    }
    \subfloat[\label{fig:turning_graph}]{%
        \includegraphics[height = 0.319\linewidth]{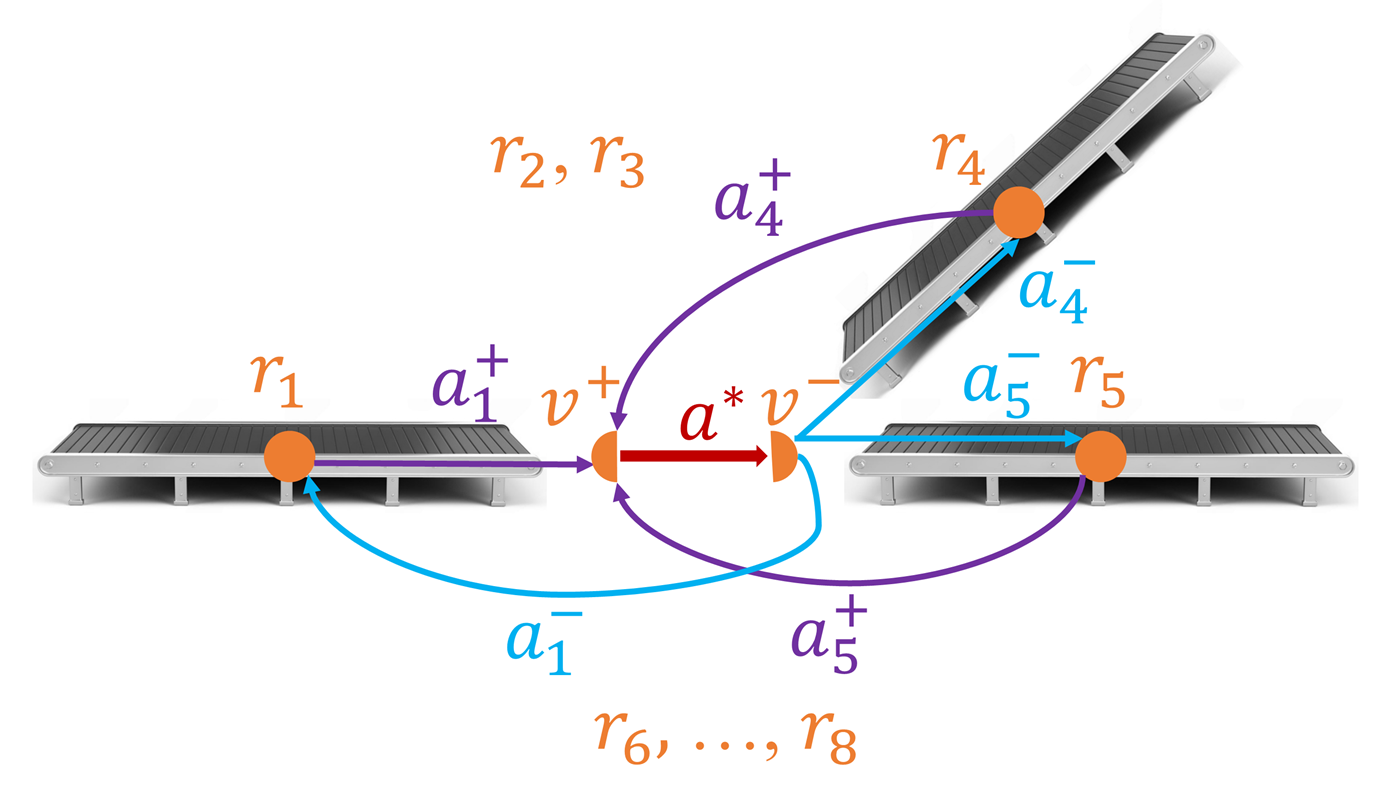}
    }
    \caption{Inline junctions (a) and multi-way/turning junctions (b) in the graph}
    \label{fig:junctions_graph}
\end{figure}

An inline junction can only connect two adjacent belt segments that share the same orientation.
Thus, for each pair of parallel segments at a grid point, we assign a new arc \(a^*\) connecting two adjacent belt segments \(r^+, r^-\), as shown in \cref{fig:inline_graph}.
We record the set of inline junction's arcs as $A_\mathrm{inline}$, and add the following constraints to \cref{eq:layout_optimization}:
%
\begin{subequations}
\label{eq:inline_junction}    
\begin{alignat}{4}
    s_{a^*} \le s_{r^+}, &&\ s_{a^*} \le s_{r^-}, &&\ \forall a^* &\in A_\mathrm{inline} \label{constr:inline_segment_control}\\
    \sum_{a^*\in \delta^+(r)} s_{a^*} \le 1, &&\sum_{a^*\in \delta^-(r)} s_{a^*} \le 1, &&\ \forall r &\in \mathcal{R}_\mathrm{belt}
\label{constr:inline_direction}
\end{alignat}
\end{subequations}
where \cref{constr:inline_segment_control} ensures an inline junction only appears when its parent and child segments are selected.
\Cref{constr:inline_direction} ensure that each segment has at most one inline junction starting from it and at most one ending at it so that 
the segments connected by an inline junction move in the same direction.

A multi-way junction can redirect among any subset of the eight segments that meet at a grid point.
We represent it with a pair of vertices \(v^+, v^-\) and an auxiliary arc \(a^*\).
In addition, we create eight new arcs $a^+_i$ connecting the neighboring segments to $v^+$ and eight new arcs $a^-_i$ point out from $v^-$, as shown in \cref{fig:turning_graph}. 
Such graph construction will ensure that the junction arc $a^*$ is selected when a flow exists between any pair among the eight segments.


A turning junction can be seen as a special case of a multi-way junction, where only two segments are connected.
We can encode this with the following constraints:
\begin{equation}
\begin{alignedat}{1}
    &\sum_{i = 1}^{8} s_{a_i^+} = 1,\ \sum_{i = 1}^{8} s_{a_i^-} = 1,\\
\end{alignedat}
\end{equation}

Finally, we assign cost terms $w_{a}$ for all the auxiliary arcs representing the conveyor belt segments and junctions.
A unit-length conveyor belt segment requires a motor to operate its belt, so its cost is the sum of the belt cost and the motor cost.
When multiple segments are connected by an inline junction, they share one motor, so their cost is the sum of the unit-length belt cost times the total length plus the cost of a motor.
Inline junctions can reduce the motor cost of a composed conveyor belt system. Thus, their costs are set to negative to encourage their adoption.
Multi-way and turning junctions have positive costs, with the multi-way junctions being more expensive due to their mechanical complexity.

\subsection{Implementing payload constraints}
Our layout optimization can be conveniently configured to respond to payload capabilities of different types of robot manipulators.
This constraint is enforced by adding 
\begin{equation}
    f_{a,i}=0, \forall a\in A_{aux}\ s.t.\ \textsc{Payload}(a) < \textsc{Weight}_i
\end{equation}
in \cref{eq:layout_optimization}, preventing any robot $a$ with insufficient payload capacity from delivering box $i$.

\section{Results}
We evaluate the proposed layout design framework in a kinematic simulator. 
The experiments were implemented in C++ and executed on a Ubuntu computer with an Intel i7-8700K CPU @ 3.7 GHz and 32 GB RAM.
We use Gurobi \cite{gurobi} as the MILP solver for layout optimization (\cref{eq:layout_optimization}).

\subsection{Comparison with Heuristic Search}\label{sect:comparison}
Our graph-based layout optimization determines all robot placement {\it jointly} with connectivity enforced by flow constraints.
An intuitive alternative way to solve this is to assign robots {\it procedurally} using a heuristic search to reach multiple destinations.

We choose the A* algorithm as a baseline and compare it with our method on the test scenes with different grid resolutions and different output locations. Only UR5e is considered in these two experiments.
The input to the A$^*$ search is the position of all input/output locations, and the output is a robot placement to deliver all boxes with the goal of minimizing the total number of UR5e assigned.
The state of the search is a set of grid position-robot pairs: \(q = \{(p, r) \in P \times \mathcal{R} \}\). 
The neighbors of a given state \(q\) include all robot types assigned on reachable grid points, which makes the branching factor quite large.
For the objective function \(f(q) = g(q) + h(q)\), the current cost \(g(q)\) considers the number of UR5e already selected in the path.

The main challenge for making A$^*$ efficient is the design of an admissible, yet tight heuristic function, 
We select a heuristic that computes the minimum number of UR5es needed to connect the farthest output location to the current reachable blob: 
\[
h(q) = \left\lceil \frac{1}{2l}\max_{i\in\{1, \ldots, N\}} \min_{p\in q} \textsc{Dist}_i(p)\right\rceil 
\]
where \(l\) denotes the maximal reach distance of UR5e and \(\textsc{Dist}_i(p)\) denotes the distance from the grid point \(p\in P\) to the output location of box \(i\).

In the following two experiments, the floor is set as an $8\,\text{m} \times 8\,\text{m}$ square, spanning from $(0, -4)$ to $(8,4)$. The input location is fixed at $(0, 0)$, while all output locations are randomly sampled within the rectangular region from $(6,-4)$ to $(8,4)$ with each pair of output locations being at least one meter apart. Additionally, a time limit of 300 seconds is imposed.


\subsubsection{Impact of Grid Resolution}\label{sect:resolution}
\begin{figure}[thpb]
    \centering
    \cornerlabel
    \subfloat[\label{fig:resolution_time}]{%
        \includegraphics[width = 0.485\linewidth]{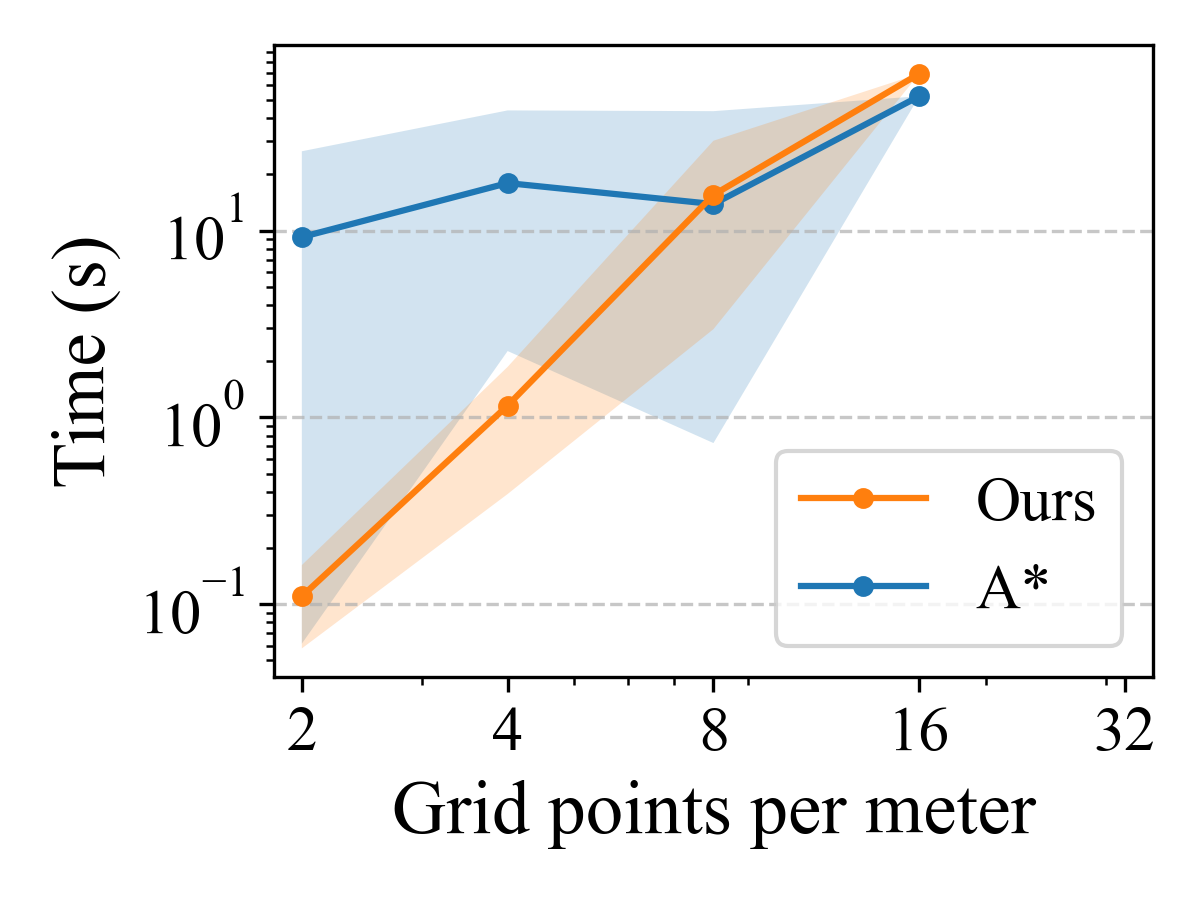}
    }
    \subfloat[\label{fig:resoluiton_success}]{%
        \includegraphics[width = 0.485\linewidth]{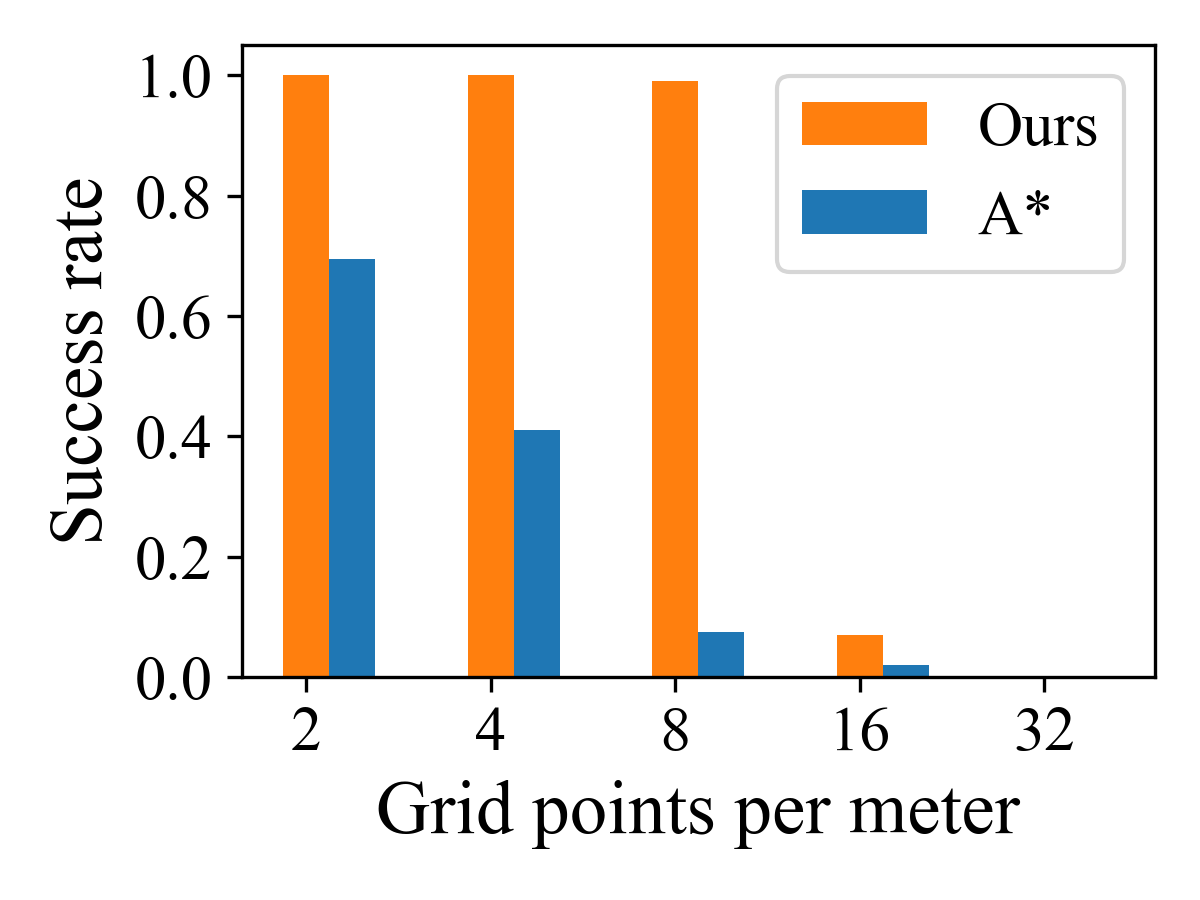}
    }
    \caption{Impact of grid resolution on running time (a) and success rate (b). The shaded area in (a) represents the running time range from the 10th percentile to the 90th percentile.}
    \label{fig:resolution}
\end{figure}

In this experiment, we evaluate the performance of our algorithm under different grid resolutions. We randomly generate 100 cases, each with two output locations, and compare the performance of our algorithm with A$^*$. The running time and the success rate are shown in \cref{fig:resolution_time} and \cref{fig:resoluiton_success}.

For consistency, \cref{fig:resolution_time} only includes cases that both methods succeed. As shown in \cref{fig:resolution_time}, our method achieves an average acceleration of 100$\times$ when the resolution is $0.5\,\text{m}$. As the number of grid points per meter grows larger, our algorithm performs at least as well as the baseline. In addition, the variation in running time for our method is smaller compared to the baseline, demonstrating greater stability against different problem settings in execution.

Additionally, \cref{fig:resoluiton_success} demonstrates that our algorithm achieves a higher success rate compared to A$^*$. Another observation, not explicitly shown in these figures, is that when the resolution is coarser than 32 grid points per meter, A$^*$ always fails due to running out of memory, while our method fails for timeout. It is only when the resolution reaches 32 grid points per meter that both methods fail due to running out of memory---highlighting the better memory efficiency of our method. 

\subsubsection{Impact of Output Locations Count}
\begin{figure}[thpb]
    \centering
    \cornerlabel
    \subfloat[\label{fig:ports_time}]{%
        \includegraphics[width = 0.485\linewidth]{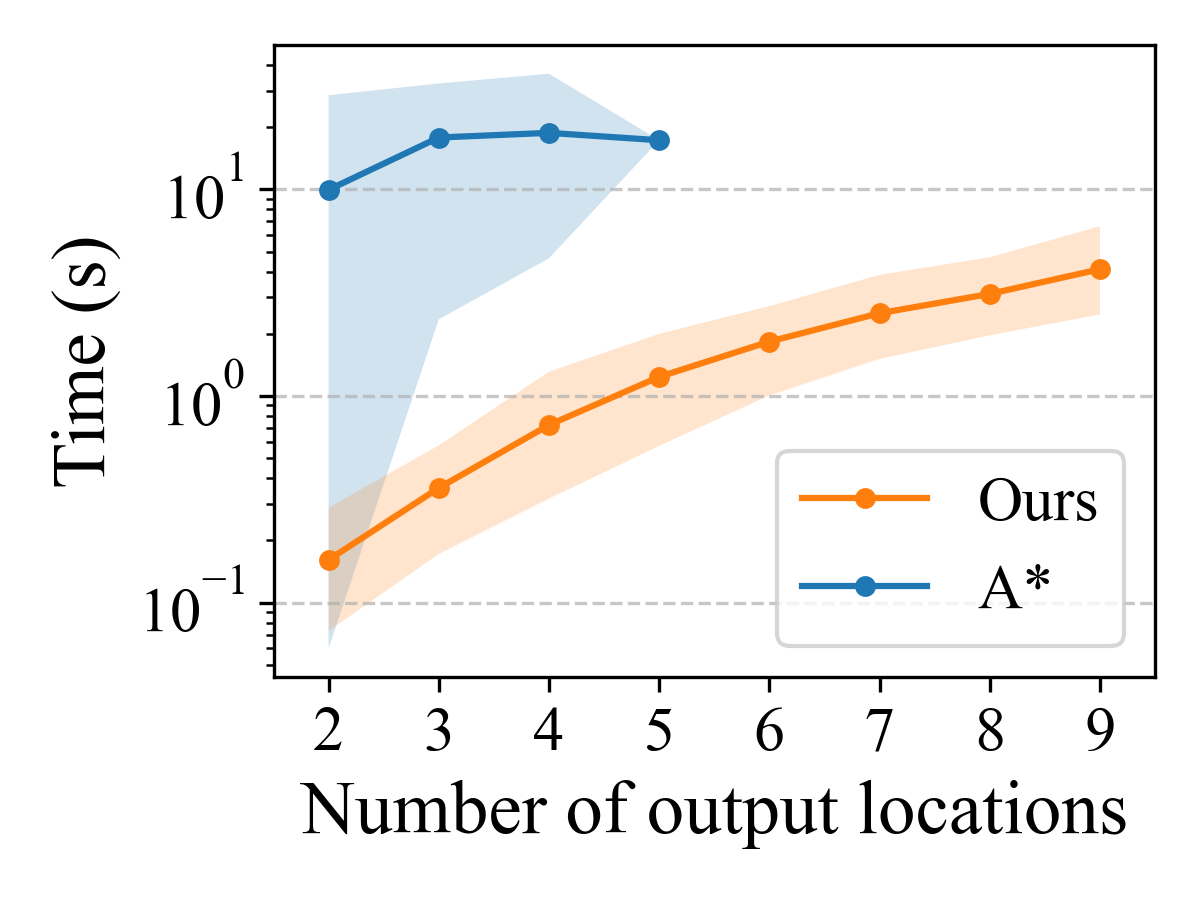}
    }
    \subfloat[\label{fig:ports_success}]{%
        \includegraphics[width = 0.485\linewidth]{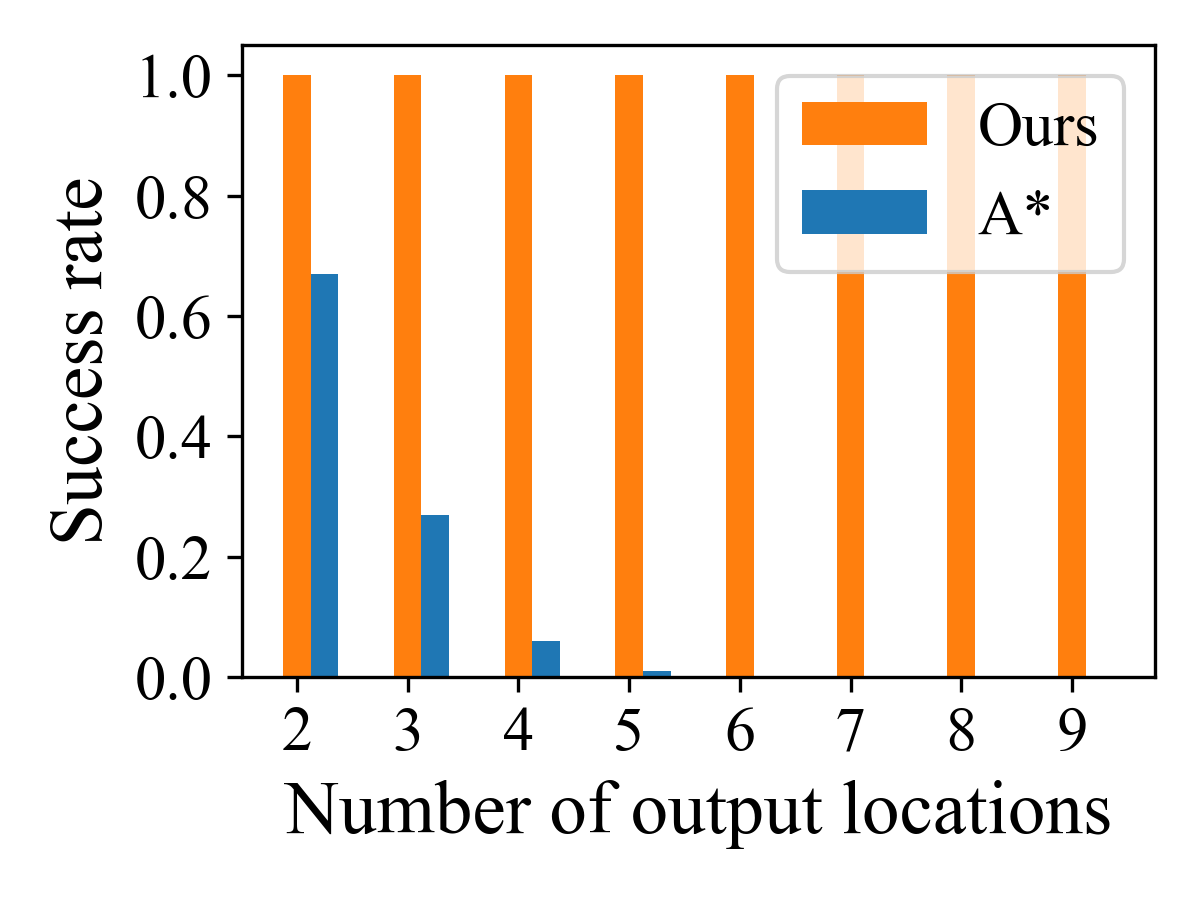}
    }
    \caption{Impact of output locations count on running time (a) and success rate (b). The shaded area in (a) represents the running time range from the 10th percentile to the 90th percentile.}
    \label{fig:ports}
\end{figure}

We fixed the resolution to $0.5\,\text{m}$ and evaluate the performance based on different numbers of output locations. For each setting, both algorithms were run 100 times. The running time and success rate are presented in \cref{fig:ports_time} and \cref{fig:ports_success}, respectively. 

As shown in \cref{fig:ports_time}, our method remains at least 10 times faster in all cases. In \cref{fig:ports_success}, our method achieves a success rate of 100\% all the time, while the success rate of A$^*$ drops sharply as the number of output locations increased, which demonstrates our algorithm's greater robustness against complex scenarios. 

\subsection{Extensions}
To demonstrate the extensions mentioned in \cref{sect:extensions}, we conduct experiments on multiple simulated scenarios. 
In all scenarios, the spacing between grid points for layout optimization is set to $0.5\,\text{m}$.
The reference grids visualized in all the figures have a one-meter spacing. The standard cost settings and running times for experiments are listed in \cref{tab:standard_costs} and \cref{tab:running_time}, respectively.

\begin{table}[thpb]
    \centering
    \caption{Standard cost settings for experiments, normalized proportional to the market price.}
    \begin{tabular}{cccccc}
        \toprule
        UR5e & IRB4600 & 
        \begin{tabular}{@{}c@{}}Belt\\(per meter)\end{tabular}&
         Motor & 
        \begin{tabular}{@{}c@{}}Multi-way\\Junction\end{tabular}&
        \begin{tabular}{@{}c@{}}Turning\\Junction\end{tabular}\\
        \midrule
        1.0 & 3.0 & 0.2 & 0.1 & 0.1 & 0.05\\
        \bottomrule
    \end{tabular}
    \label{tab:standard_costs}
\end{table}

\begin{table*}[thpb]
    \centering
    \caption{MILP parameters and running time of each experiment.}
    \label{tab:running_time}
    \begin{tabular}{cc@{\ \ }c@{\ \ }cc@{\ \ }c@{\ \ }c}
        \toprule
        & \multicolumn{3}{c}{MILP parameters} & 
        \multicolumn{3}{c}{\multirow{2}{*}{Running Time (in seconds)}} \\
        \cmidrule(lr){2-4}
        Experiment & \multicolumn{2}{c}{Number of variables} & 
        \multirow{2.4}{*}{\begin{tabular}{@{}c@{}}Number of\\ constraints\end{tabular}} & & & \\
        \cmidrule(lr){2-3} \cmidrule(lr){5-7}
        & Continuous & Binary & & Layout Optimization & Task Scheduling & Motion Planning\\
        \midrule
        An illustrative case & 30724 & 15362 & 38654 & 10.73 & 0.001 & 5.09\\
        \midrule
        \multirow{2}{*}{\begin{tabular}{@{}r@{\ }l@{}} 
            Impact of junction costs & (Cheap junctions) \\
            Impact of junction costs & (Expensive junctions) \\
        \end{tabular}}
        & 95032 & 23758 & 112162 & 86.07 & 0.001 & 62.79\\
        & 95032 & 23758 & 112162 & 10.33 & 0.001 & 45.32\\
        \midrule
        \multirow{2}{*}{\begin{tabular}{@{}r@{\ }l@{}} 
            Handling payload constraints & (2 heavy \& 1 light) \\
            Handling payload constraints & (1 heavy \& 2 light) 
        \end{tabular}} & 974832 & 324944 & 1005896 & 30.84 & 0.001 & 2.28\\
        & 974832 & 324944 & 1005539 & 32.31 & 0.001 & 5.23\\
        \bottomrule
    \end{tabular}
\end{table*}

\subsubsection{Conveyor belts}
\begin{figure}[thpb]
    \centering
    \includegraphics[width = 0.6\linewidth]{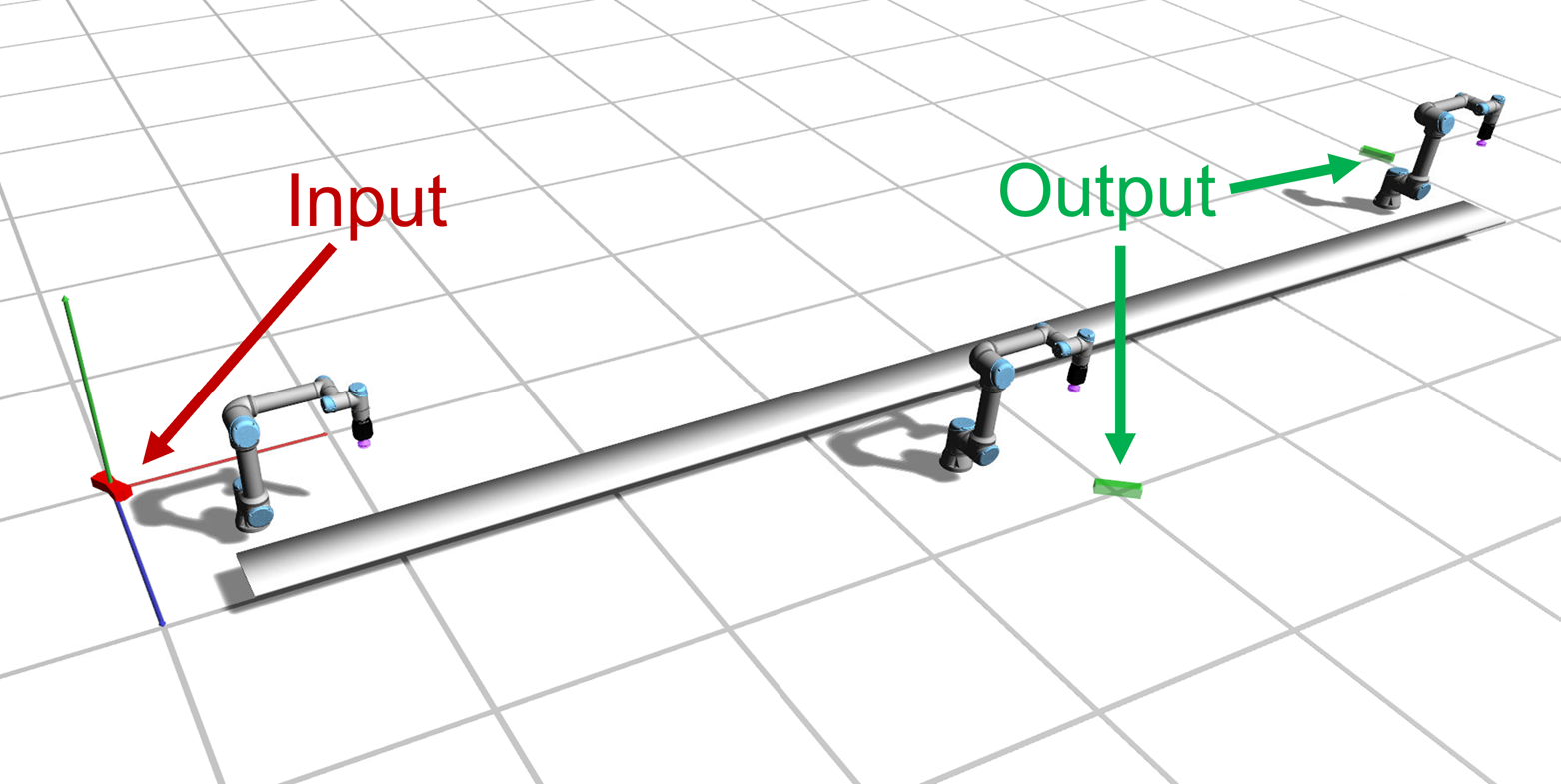}
    \caption{Optimal layout for (a) higher and (b) lower motor costs.}
    \label{fig:case3}
\end{figure}

In the following experiments, we include conveyor belts and UR5e arm in the available robot set.
\paragraph{An illustrative case}
We first present an interesting optimal layout, as shown in \cref{fig:case3}. With two distant output locations, our algorithm decides that using a long straight conveyor belt to connect them is more cost-efficient than splitting it up, as the input location and the two output locations are nearly on a straight line. Moreover, two robots are assigned to different sides of the belt to unload the two boxes.

\begin{figure}[thpb]
    \centering
    \cornerlabel
    \subfloat[\label{fig:case2_1}]{%
        \includegraphics[width = 0.485\linewidth]{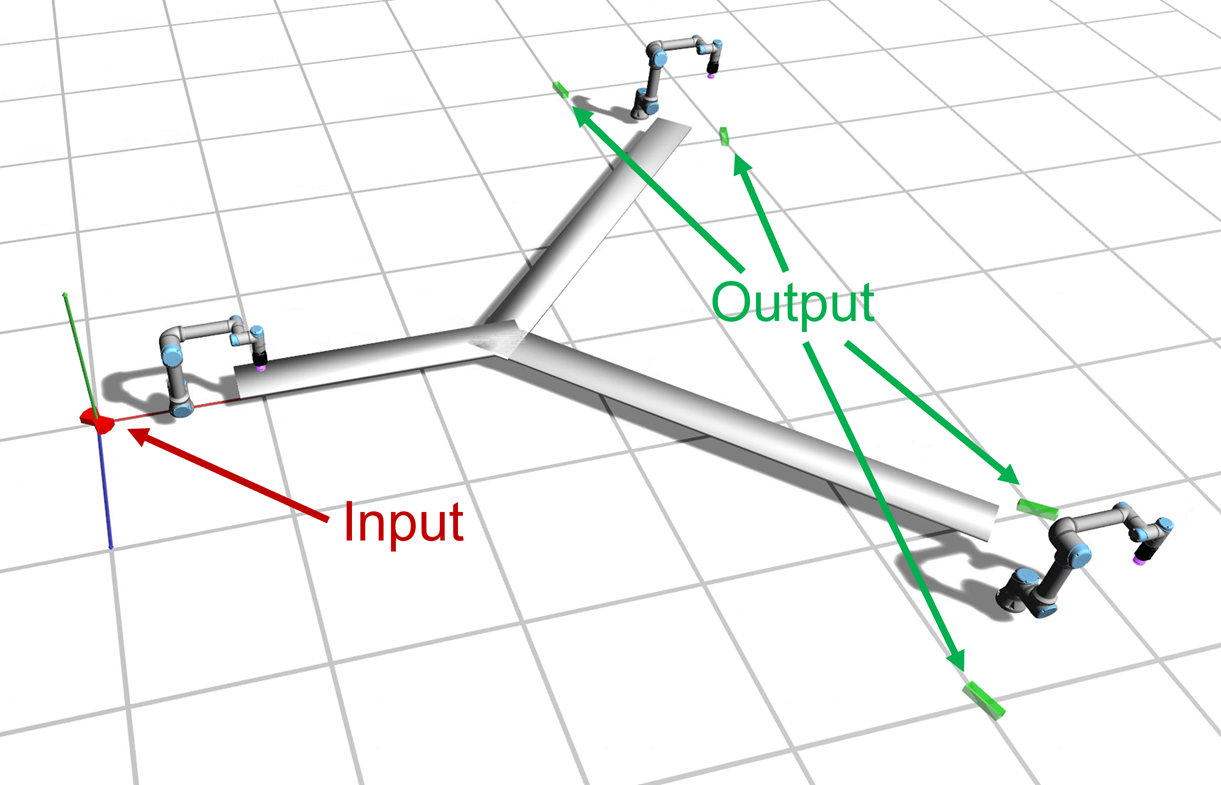}
    }
    \subfloat[\label{fig:case2_2}]{%
        \includegraphics[width = 0.485\linewidth]{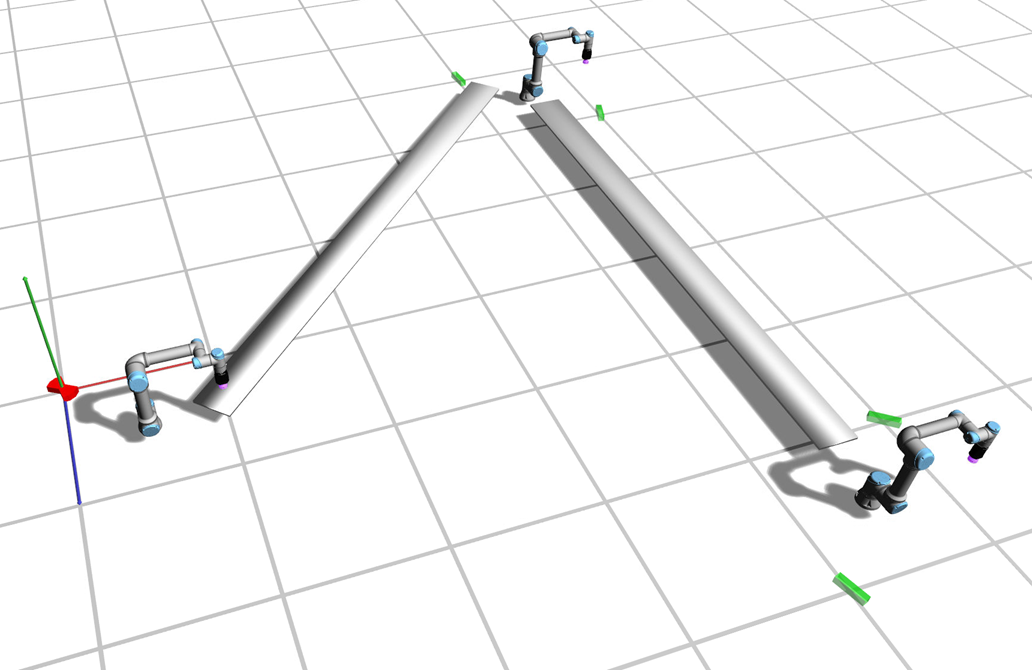}
    }
    \caption{Optimal layout for (a) lower and (b) higher junction costs.}
    \label{fig:case2}
\end{figure}

\paragraph{Impact of junction costs}
In this experiment, after including conveyor belts, we showcase how the layout optimization responds to different junction cost configurations.
We normalize the price of a UR5e robot arm to \(1.0\) and set the price of conveyor belts and junctions according to \cref{tab:standard_costs}.
In \cref{fig:case2_1}, we show the optimal layout that automatically groups the four outputs into two pairs, and assign each pair with a robot arm to unload.
In the center, it chooses a multi-way junction to split the flow to different unloading robot arms. 

In contrast, after quadrupling the price of multi-way and turning junctions, the optimal layout in \cref{fig:case2_2} uses two long belts with inline junctions instead of the more expensive multi-way junction.

\paragraph{Optimal layout for different output locations setting}

\begin{figure}[thpb]
    \centering
    \cornerlabel
    \subfloat[\label{fig:scalability1_1}]{%
        \includegraphics[width = 0.22\linewidth]{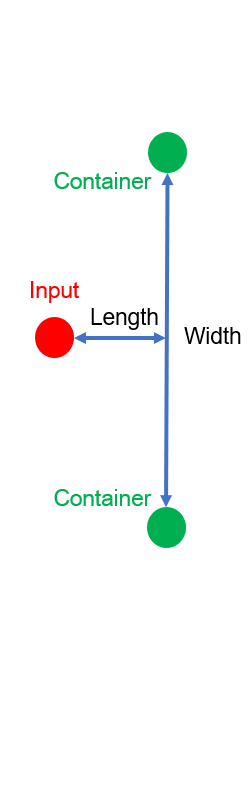}
    }
    \subfloat[\label{fig:scalability1_2}]{%
        \includegraphics[width = 0.74\linewidth]{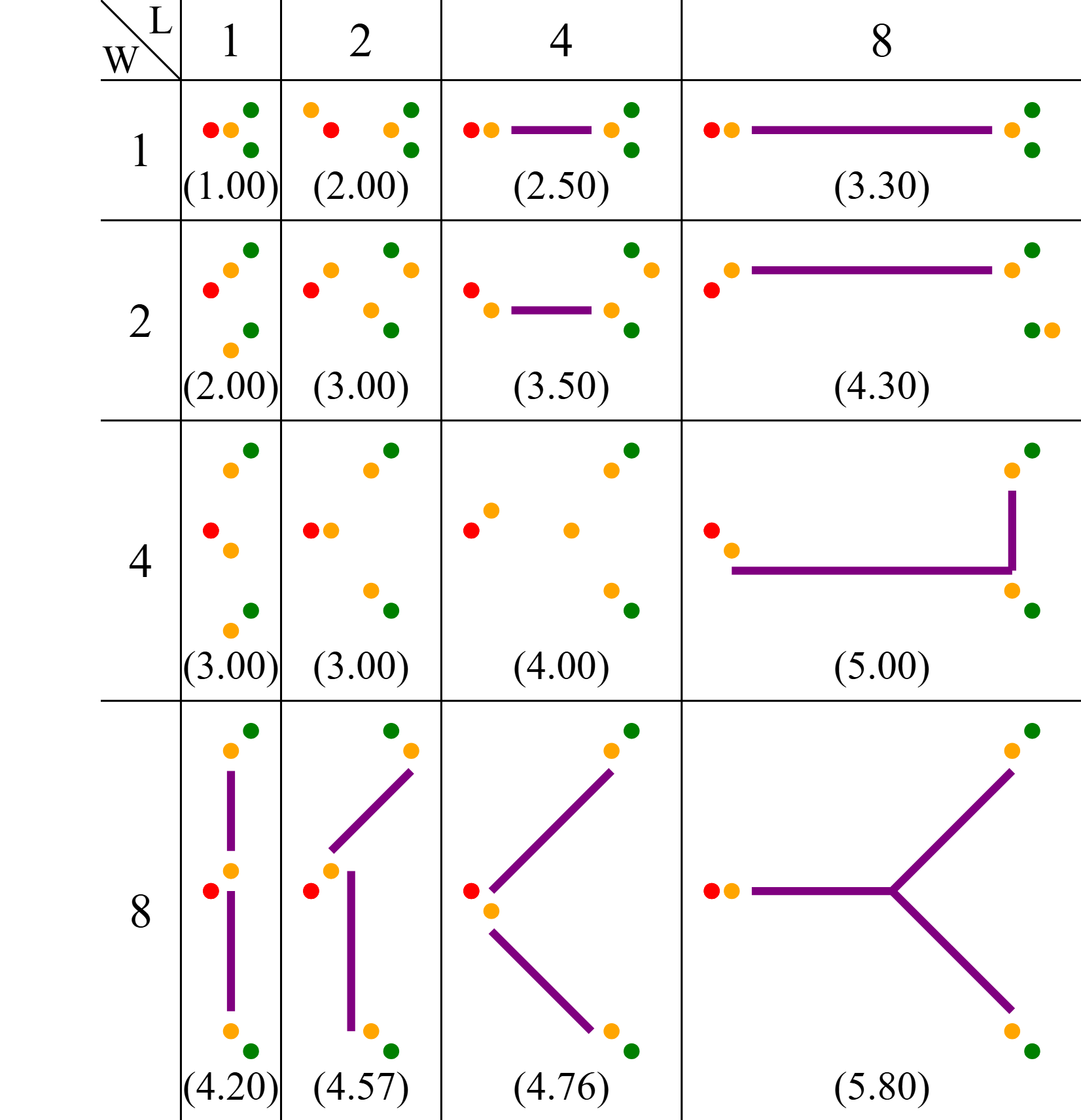}
    }
    \caption{Test scenes (a) from the scalability analysis and results (b) under different lengths and widths, showing optimal layouts and corresponding costs (in parentheses)}
    \label{fig:scalability}
\end{figure}

To demonstrate the robustness of our method, we set up a series of scenarios to test it.
As shown in \cref{fig:scalability1_1}, there are two output locations set symmetrically to the input. By adjusting the length and the width, we can change the grid size and, thus, the input size. The larger the length or width, the more grid points and corresponding robot candidates are considered in the optimization process. 

Our goal is to allocate UR5e and conveyor belts to solve the tasks with varying length and width.
\cref{fig:scalability1_2} shows the optimal layouts, where orange dots stand for UR5e and purple segments stand for conveyor belts. An interesting case is the $4\,\text{m} \times 4\,\text{m}$ one, where no conveyor belts appear, unlike other 4-meter-length cases. The algorithm decides that one robot arm is more cost-efficient than multiple short belts connected with a junction.

\subsubsection{Handling payload constraints}
Besides conveyor belts, our layout optimization can be further configured to respond to robot manipulators with different payload capabilities. In this experiment, we include IRB4600 into the available robot set.

\begin{figure}[thpb]
    \centering
    \cornerlabel
    \subfloat[\label{fig:case4_1}]{%
        \includegraphics[width = 0.485\linewidth]{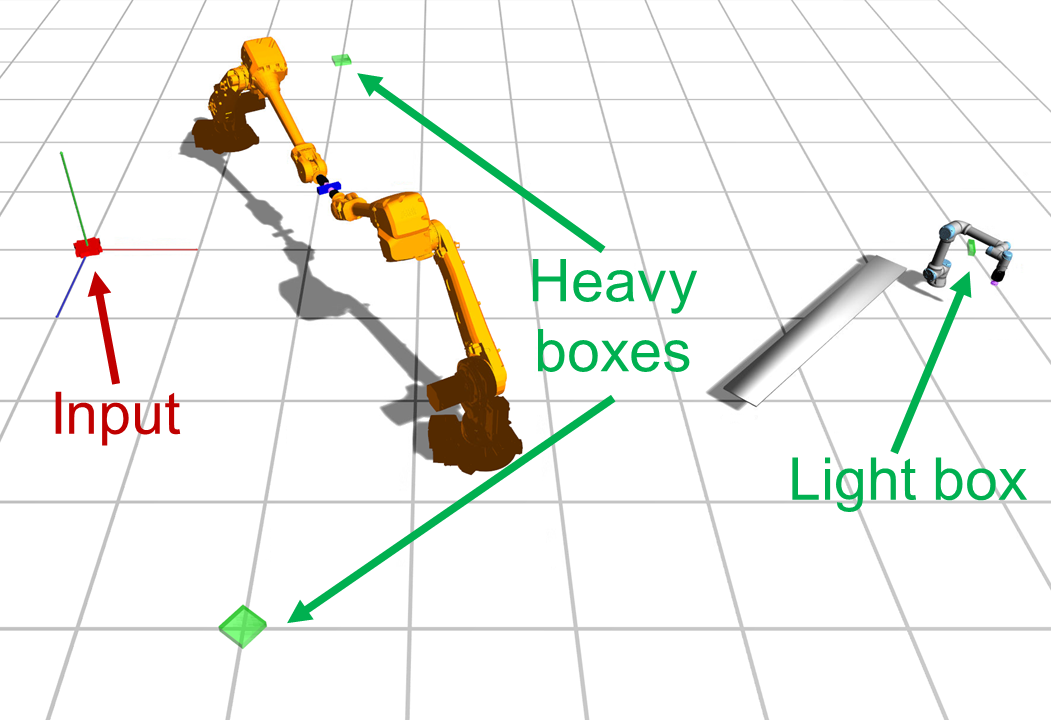}
    }
    \subfloat[\label{fig:case4_2}]{%
        \includegraphics[width = 0.485\linewidth]{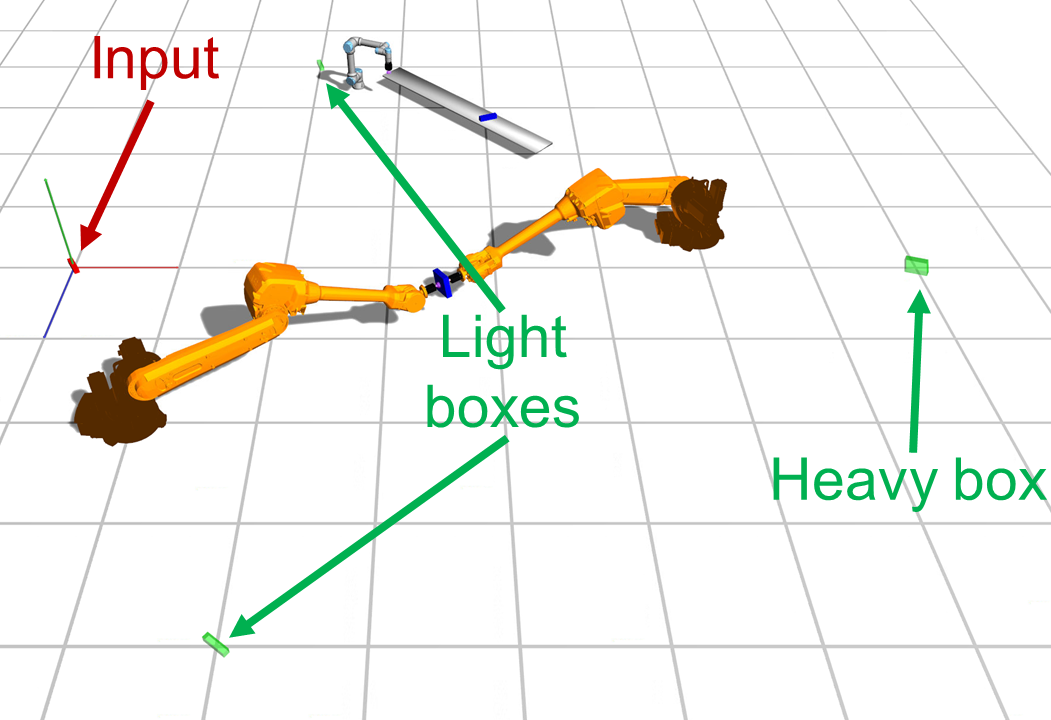}
    }
    \\
    \subfloat[\label{fig:case4_1c}]{%
        \includegraphics[width = 0.485\linewidth]{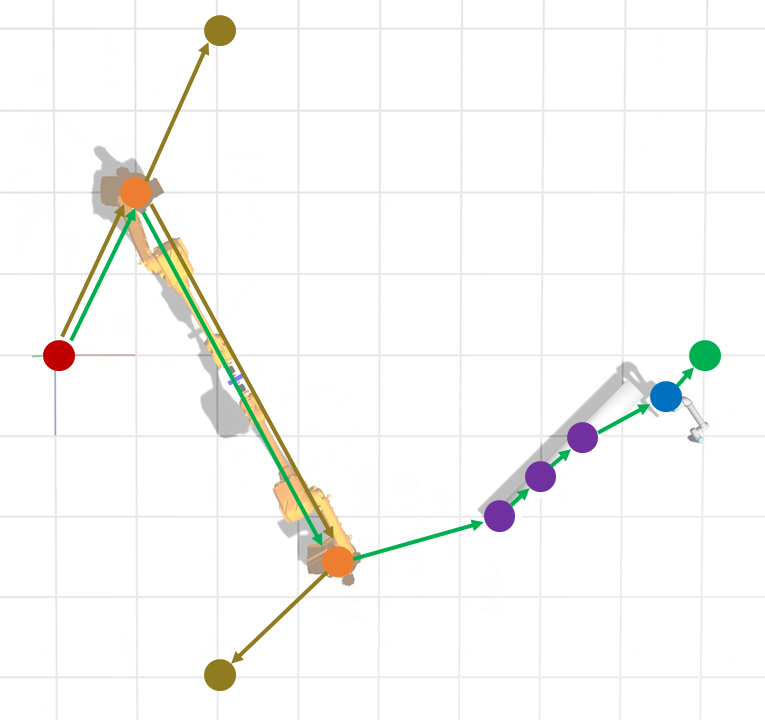}
    }
    \subfloat[\label{fig:case4_2c}]{%
        \includegraphics[width = 0.485\linewidth]{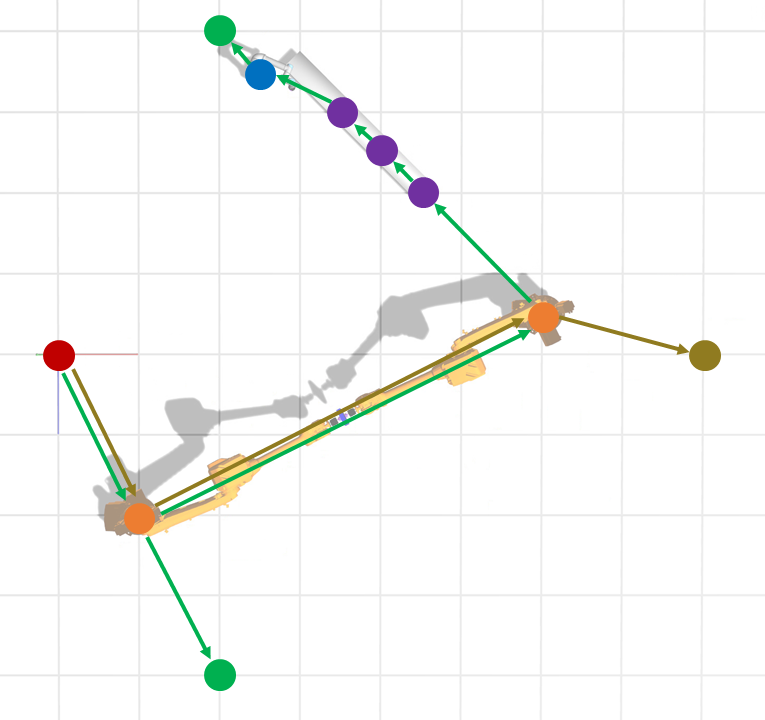}
    }
    \caption{Optimal Layout (a,b) and box delivery paths (c,d) when handling a scenario with two heavy boxes and one light box (a,c) and a scenario with one heavy box and two light boxes (b,d). Three types of robots --- UR5e (blue), IRB4600 (orange), and conveyor belts (purple) --- are employed.}
    \label{fig:case4}
\end{figure}

Under the same input and output locations, 
\Cref{fig:case4} shows the generated layouts corresponding to the same output locations, but different box weight distributions.

From \cref{fig:case4}, we can notice IRB4600s are positioned closer to heavy boxes since each heavy box requires an IRB4600 to carry and place down. In contrast, each light box can be handled with a combination of UR5es and conveyor belts to save budget.
Also, conveyor belts appear less frequently than in previous experiments since the large reachable area of the IRB4600 reduces the reliance on conveyor belts.

\section{Discussion and Conclusion} \label{sect:discussion}
We have presented a new graph flow-based formulation to optimize the layout of a multi-robot box delivery system with multiple destinations.
Our method can compute the globally optimal layout to minimize hardware budget in a memory efficient way.
Our method can be extended to incorporate different manipulators and conveyor belt, payload constraints, and cost assignment.
To our best knowledge, this is the first work to address this critical yet under-explored problem of optimizing layout for multiple robot manipulators.


\paragraph*{Limitations and future work}
First, our optimization-based method can only handle cases with a resolution coarser than 32 grid points per meter in \cref{sect:comparison} due to RAM limitations. Separation oracle, a well-known concept in cutting-plane methods for linear programming \cite{grotschel2012geometric}, can be implemented to reduce RAM usage. As we formulate the subgraph extraction problem into a MILP, it is straightforward to develop a separation oracle to judge the feasibility of a given MILP solution by testing it in the original graph problem. Such cutting-plane methods are implemented in most off-the-shelf MILP solvers. Additionally, a well-developed heuristic can be introduced into the MILP solver to accelerate the solving process.

Second, we focus on the layout optimization and do not address many crucial aspects for practical logistic systems, such as dynamic task allocation \cite{harris2022fc} and reactive control strategies \cite{toussaint2022sequence} to respond to changing box influx schedule and uncertainties, and vision system integration.
Integrating our layout generation with these aspects is an important future direction towards a flexible and robust logistic system.


Thirdly, our layout generation optimizes for hardware budget, without considering throughput, another important objective when designing practical logistic systems.
However, estimation of throughput relies on time-optimal motion planning for multiple high-dof manipulators, which is a challenging problem itself and known to have multiple local-minimas.
Future work could explore ways to formally define a throughput score and develop a way to approximate its computation, after which, together with our approach, a multi-objective optimization could be used to find the Pareto front balancing budget and throughput. 
If an optimal solution for plotting the monotonically increasing budget-throughput curve is eventually developed, this paper could still serve as a reference for the bottom-left point of the curve.

Finally, we focus on planning layout for stationary, fully actuated robots on a pre-discretized floor grid.
While this is a common setting for today's logistic systems, a promising direction is to relax the discrete grid assumption and allow continuous placement for increasingly capable wheel-based or legged mobile robots. 

As robots are required to collaborate to achieve increasingly complex tasks, we believe this work presents a step towards a future where robots not only have intelligence for their individual skills, but also have collective intelligence to actively shape their shared workspace to improve system-wide efficiency.


\addtolength{\textheight}{-12cm}   









\bibliographystyle{IEEEtran}
\bibliography{IEEEabrv.bib, references.bib}

\end{document}